\documentclass[letterpaper,onefignum,onetabnum]{siamonline190516}

\usepackage{amssymb}
\usepackage{amsopn}
\usepackage{mathtools}
\usepackage{amsfonts}
\usepackage{relsize}
\usepackage{subfig}
\usepackage{stackengine}
\usepackage{lipsum}
\usepackage{graphicx}
\usepackage{epstopdf}
\usepackage{bm}
\usepackage{algorithmic}
\ifpdf
  \DeclareGraphicsExtensions{.eps,.pdf,.png,.jpg}
\else
  \DeclareGraphicsExtensions{.eps}
\fi

\usepackage{enumitem}
\setlist[enumerate]{leftmargin=.5in}
\setlist[itemize]{leftmargin=.5in}

\newsiamremark{remark}{Remark}
\newsiamremark{hypothesis}{Hypothesis}
\crefname{hypothesis}{Hypothesis}{Hypotheses}
\newsiamthm{claim}{Claim}

\headers{Tensor Regression Using Low-rank and Sparse Tucker Decompositions}{T.~Ahmed, H.~Raja, and W.~U.~Bajwa}

\title{Tensor Regression Using Low-rank and Sparse Tucker Decompositions%
\thanks{Date of last revision:~\today.%
\funding{This work was supported in part by the National Science Foundation (NSF) under awards CCF-1453073 and CCF-1910110, and by the Army Research Office (ARO) under award W911NF-17-1-0546.}}
}

\author{Talal~Ahmed\thanks{Department of Electrical and Computer Engineering, Rutgers, The State University of New Jersey, Piscataway, NJ 08854, USA (\email{talal.ahmed@rutgers.edu}, \email{waheed.bajwa@rutgers.edu}).}%
\and Haroon~Raja\thanks{Department of Electrical Engineering and Computer Science, University of Michigan, Ann Arbor, MI 48109, USA (\email{hraja@umich.edu}).}%
\and Waheed~U.~Bajwa\footnotemark[2]}

\ifpdf
\hypersetup{
  pdftitle={Tensor Regression Using Low-rank and Sparse Tucker Decompositions},
  pdfauthor={T. Ahmed, H. Raja, and W. U. Bajwa}
}
\fi

\mathchardef\mhyphen="2D

\newcommand{\covering}{\ensuremath{\Psi}}
\newcommand{\M}{\ensuremath{M}}
\newcommand{\Mset}{\ensuremath{\mathcal{M}}}
\newcommand{\Classzero}{\ensuremath{\mathcal{C}}}
\newcommand{\Class}{\ensuremath{\mathcal{G}_{\rank,\sparsity, \corebound }}}

\newcommand{\X}{\ensuremath{ \mathbf{X} }}
\newcommand{\W}{\ensuremath{ \mathbf{B} }}
\newcommand{\B}{\ensuremath{ \mathbf{B} }}
\newcommand{\Vz}{\ensuremath{ V_\mathbf{Z}}}

\newcommand{\Z}{\ensuremath{ \mathbf{Z}}}
\newcommand{\Ga}{\ensuremath{\mathbf{G}_a}}
\newcommand{\Gb}{\ensuremath{ \mathbf{G}_b}}
\newcommand{\Sa}{\ensuremath{ \mathbf{S}_a}}
\newcommand{\Sb}{\ensuremath{ \mathbf{S}_b}}
\newcommand{\U}{\ensuremath{ U}}
\newcommand{\termj}{\ensuremath{b_j}}

\newcommand{\Walgo}{\ensuremath{\W}^{*}}
\newcommand{\Wtildek}{\ensuremath{\widetilde{\W}^{k}}}
\newcommand{\Wtilde}{\ensuremath{\widetilde{\W}}}
\newcommand{\What}{\ensuremath{\widehat{\W}}}
\newcommand{\Wk}{\ensuremath{ {\W} ^{k}}}

\newcommand{\Wkplus}{\ensuremath{ {\W} ^{k+1}}}

\newcommand{\corebound}{\tau}

\newcommand{\Zvec}{\ensuremath{\mathbf{z}}}

\newcommand{\A}{\ensuremath{\mathcal{X}}}
\newcommand{\Aconjugate}{\ensuremath{\mathcal{X}^{*}}}
\newcommand{\Loss}{\ensuremath{\mathcal{L}}}
\newcommand{\y}{\ensuremath{\mathbf{y}}}
\newcommand{\sparsity}{\ensuremath{\barbelow{s}}}
\newcommand{\rank}{\ensuremath{\barbelow{r}}}
\newcommand{\deltariptwo}{\ensuremath{\delta_{2\rank, \sparsity, 2\corebound}}}
\newcommand{\deltarip}{\ensuremath{\delta_{\rank, \sparsity, \corebound}}}
\newcommand{\g}{\ensuremath{g}}
\newcommand{\Score}{\ensuremath{\mathbf{S}}}
\newcommand{\Scorehat}{\ensuremath{\widehat{\mathbf{S}}}}
\newcommand{\Hsr}{\ensuremath{\mathcal{H}}}

\newcommand{\noise}{\boldsymbol{\eta}}

\newcommand{\dist}{\ensuremath{h}}
\newcommand{\normed}{\ensuremath{\|\, . \,\|}}

\newcommand\eqa{\stackrel{\mathclap{\normalfont\mbox{(a)}}}{=}}
\newcommand\eqb{\stackrel{\mathclap{\normalfont\mbox{(b)}}}{=}}

\newcommand\leqa{\stackrel{\mathclap{\normalfont\mbox{(a)}}}{\leq}}
\newcommand\leqb{\stackrel{\mathclap{\normalfont\mbox{(b)}}}{\leq}}
\newcommand\leqc{\stackrel{\mathclap{\normalfont\mbox{(c)}}}{\leq}}
\newcommand\leqd{\stackrel{\mathclap{\normalfont\mbox{(d)}}}{\leq}}
\newcommand\leqe{\stackrel{\mathclap{\normalfont\mbox{(e)}}}{\leq}}
\newcommand\leqf{\stackrel{\mathclap{\normalfont\mbox{(f)}}}{\leq}}
\newcommand\leqg{\stackrel{\mathclap{\normalfont\mbox{(g)}}}{\leq}}
\newcommand\leqh{\stackrel{\mathclap{\normalfont\mbox{(h)}}}{\leq}}
\newcommand\leqi{\stackrel{\mathclap{\normalfont\mbox{(i)}}}{\leq}}
\newcommand\leqj{\stackrel{\mathclap{\normalfont\mbox{(j)}}}{\leq}}
\newcommand\leqk{\stackrel{\mathclap{\normalfont\mbox{(k)}}}{\leq}}

\newcommand\barbelow[1]{\stackunder[1.2pt]{$#1$}{\rule{.8ex}{.075ex}}}

\newcommand{\revise}[1]{#1}

\begin{document}

\maketitle

\begin{abstract}
This paper studies a tensor-structured linear regression model with a scalar response variable and tensor-structured predictors, such that the regression parameters form a tensor of order $d$ (i.e., a $d$-fold multiway array) in $\mathbb{R}^{n_1 \times n_2 \times \cdots \times n_d}$. In particular, it focuses on the task of estimating the regression tensor from $m$ realizations of the response variable and the predictors where $m\ll n = \prod \nolimits_{i} n_i$.  Despite the seeming ill-posedness of this estimation problem, it can still be solved if the parameter tensor belongs to the space of sparse, low Tucker-rank tensors. Accordingly, the estimation procedure is posed as a non-convex optimization program over the space of sparse, low Tucker-rank tensors, and a tensor variant of projected gradient descent is proposed to solve the resulting non-convex problem. In addition, mathematical guarantees are provided that establish the proposed method \revise{linearly converges to an appropriate solution under a certain set of conditions}. Further, an upper bound on sample complexity of tensor parameter estimation for the model under consideration is characterized for the special case when the individual (scalar) predictors independently draw values from a sub-Gaussian distribution. The sample complexity bound is shown to have a polylogarithmic dependence on $\bar{n} = \max \big\{n_i: i\in \{1,2,\ldots,d \} \big\}$ and, orderwise, it matches the bound one can obtain from a heuristic parameter counting argument. Finally, numerical experiments demonstrate the efficacy of the proposed tensor model and estimation method on a synthetic dataset and a collection of neuroimaging datasets pertaining to attention deficit hyperactivity disorder. Specifically, the proposed method exhibits better sample complexities on both synthetic and real datasets, demonstrating the usefulness of the model and the method in settings where $n \gg m$.
\end{abstract}



\section{Introduction}\label{sec:Introduction}
Many modern data science problems involve learning a high-dimensional regression model, where the number of predictors is much larger than the number of samples. We focus on \emph{tensor-structured} regression models, where the predictors appear naturally in the form of a tensor. Such regression models find applications within hyperspectral imaging~\cite{landgrebe2002hyperspectral, bioucas2012hyperspectral}, climatology~\cite{jolliffe2012forecast}, neuroscience~\cite{acar2007multiway, lazar2008statistical}, sentiment analysis~\cite{pang2008opinion}, and computer vision~\cite{fu2010age}. In this work, specifically, we consider a \emph{linear} tensor-structured regression model with response variable $y\in\mathbb{R}$, tensor (multiway array) of predictors $\X \in \mathbb{R}^{n_1 \times n_2 \times \cdots \times n_d}$, tensor of regression parameters $\B \in \mathbb{R}^{n_1 \times n_2 \times \cdots \times n_d}$, and noise $\eta \in \mathbb{R}$ such that $y = \langle \X, \B \rangle + \eta \text{,}$ where $d\in\mathbb{Z}^{+}$ and $\langle \cdot , \cdot \rangle$ denotes the canonical inner product. Among the various applications of this model, a major one appears in neuroimaging data analysis, where the voxels (predictors) in a brain image naturally appear in the form of a tensor and the associated disease outcome (response) appears as a scalar variable \cite{lindquist2008statistical, lazar2008statistical, hinrichs2009spatially, ryali2010sparse}.

Mathematically, let us define $\{ \X_i \}_{i=1}^{m}$, $\{ y_i \}_{i=1}^{m}$, and $\{ \eta_i \}_{i=1}^{m}$ to be the realizations of $\X$, $y$, and $\eta$, respectively, where $m$ refers to the number of observations/measurements such that $m \lll  n:= \prod \nolimits_{i} n_i$. Then, the realizations of the linear regression model can be expressed as
\begin{align}\label{eq:System1}
y_i = \langle \X_i , \B \rangle + \eta_i \text{,}   \;\;\;  i\in \{1,2,\ldots,m\} \text{.}
\end{align}
\noindent In this paper, given $\{ \X_i \}_{i=1}^{m}$ and $\{ y_i \}_{i=1}^{m}$, we focus on the task of learning the regression model in \eqref{eq:System1}, which is equivalent to estimating $\B$. Since we are considering the high-dimensional setting of $m \lll n$ in this work, the learning task is ill-posed without imposition of additional constraints on the parameter tensor $\B$. We now discuss how this challenge has been addressed in prior work.

\subsection{Relationship to Prior Work}
One simple approach to estimating $\B$ is to vectorize the regression tensor $\B$ and the realizations $\{ \X_i \}_{i=1}^{m}$ of the predictor tensor such that the model in \eqref{eq:System1} can equivalently be expressed as $y_i = \langle \text{vec}(\X_i) , \text{vec}(\B) \rangle + \eta_i$, where vec$(\cdot)$ denotes the vectorization procedure. Since this reduces the original model to a vector-valued regression model, any of the traditional sparsity promoting techniques in the literature---such as forward selection/matching pursuit \cite{james2013introduction}, least absolute shrinkage and selection operator (LASSO) \cite{tibshirani1996regression}, elastic net \cite{zou2005regularization}, adaptive LASSO \cite{zou2006adaptive}, and Dantzig selector \cite{candes2007dantzig}---can be employed for estimating $\mathbf{b} := $vec($\B$) $\in \mathbb{R}^{n}$. However, a potential drawback of the vectorization operation is that the spatial correlation structure in tensor data might be lost, and a natural question is if we can explicitly exploit this structure for learning $\B$.

Among the various notions of tensor decompositions that capture spatial relationships among entries of a tensor, a popular decomposition is the Tucker decomposition~\cite{kolda2009tensor, sidiropoulos2017tensor}. Specifically, the concept of low Tucker rank, which is the notion of rank associated with Tucker decomposition, has been successfully imposed on the regression tensor $\B$ for sample-efficient learning of tensor-structured regression models~\cite{gandy2011tensor, tomioka2011statistical, rauhut2017}. Some early convex approaches for estimating $\B$ in this regard were based on minimization of the sum of nuclear norms of matricizations of tensor $\B$ in each mode \cite{liu2009tensor, gandy2011tensor, tomioka2011statistical, mu2014square}. To undertsand the sample complexity of such learning methods, consider the special case where the $d\mhyphen$tuple $(r,r,\ldots,r)$ is the Tucker rank of $\B$ and the entries in $\X_i$ independently draw values from a Gaussian distribution for $i\in \{1,2,\ldots,m\}$. Under this special case, it was shown that convex approaches based on sum of nuclear norm minimization require $\Omega(r {\bar{n}}^{(d-1)})$ samples for estimating $\B$~\cite{mu2014square}, where $\bar{n} := \max \big\{n_i: i\in \{1,2,\ldots,d \} \big\}$. Since the number of degrees of freedom in $\B$ are on the order of $r^d + \bar{n} r d$ in this case, such sample complexity bound is clearly sub-optimal. Thus, more recently, focus has shifted to solving \emph{non-convex} formulations of the learning problem for various tensor-valued regression models, in the hope of achieving better sample complexity~\cite{yu2016learning, chen2016non, rauhut2017}. In one such recent work that studies the imposition of low Tucker rank on $\B$ \cite{rauhut2017}, it was shown that $\B$ can be \revise{learned} using $\mathcal{O}\big( (r^d + \bar{n} r d) \log d \big)$ observations, which is order optimal up to a logarithmic factor.

Although the imposition of low Tucker rank on $\B$ allows for efficient learning, the sample complexity requirement of $\mathcal{O}\big( (r^d + \bar{n} r d) \log d \big)$ poses a linear dependence on $\bar{n}$, where this linear dependence can easily become prohibitive in many application domains. For example, consider a case from neuroimaging data analysis, where a typical MRI image has size $256 \times 256 \times 256$ with $r=3$ and $d=3$~\cite{zhou2013tensor}. Clearly, $\bar{n} \gg r$ and $\bar{n} \gg d$ in this case, and question arises if we can tighten the aforementioned sample complexity bound. This goal cannot be achieved with the imposition of low Tucker rank alone on $\B$, since the degrees of freedom in $\B$, in this case, scale linearly with $\bar{n}$. Another challenge with the imposition of low Tucker rank on $\B$ is that the resulting regression model does not encompass the typical situation where the response depends on only a few of the (scalar) predictors in the model (i.e., the sparsity assumption). In this work, we address both these challenges, simultaneously, by studying the imposition of multiple structures on $\B$, as explained next.

\subsection{Our Contributions}
We study the regression model in \eqref{eq:System1} under the assumption that ($i$) the regression tensor $\B$ has \emph{low} Tucker rank (to be made precise later) and ($ii$) the factor matrices corresponding to the Tucker decomposition of $\B$ are sparse. This simultaneous imposition of structure on $\B$ allows us to address both of the aforementioned challenges. First, the imposed sparse, low-rank structure massively reduces the number of degrees of freedom in $\B$, which helps get rid of the linear dependence of sample complexity on $\bar{n}$. Second, the imposition of sparsity on the factor matrices induces sparsity in the regression tensor $\B$, which reflects the a priori belief that the response variable typically does not depend on all the (scalar) predictors, and facilitates model interpretability. Note that this simultaneous tensor structure is reminiscent of the notion of sparse PCA from matrix decomposition literature~\cite{zou2006sparse}.

From a computational perspective, we formulate the problem of learning the sparse, low Tucker-rank $\B$ as a non-convex problem, and we propose a projected gradient descent-based method to solve it. Furthermore, in our theoretical analysis, we show that the proposed computational procedure---under a certain restricted isometry assumption on realizations of the predictor tensor---\revise{converges linearly to an approximately correct solution}. In contrast, prior works that study recovery of simultaneously structured $\B$ either ($i$) formulate a convex problem for learning the parameter tensor~\cite{raskutti2015convex}, or ($ii$) impose a sparse, low CP-rank structure on $\B$~\cite{hao2018sparse, he2018boosted}, where~\cite{hao2018sparse} imposes a certain cubic structure on realizations of the predictor tensor and~\cite{he2018boosted} lacks sample complexity guarantees.

We also evaluate the introduced restricted isometry condition for the case of independentally and identically distributed sub-Gaussian (tensor-structured) predictors, and in the process, we characterize the sample complexity of parameter estimation for the case of sparse, low Tucker-rank regression tensor. We show that our sample complexity bound has only a polylogarithmic dependence on $\bar{n} := \max \big\{n_i: i\in \{1,2,\ldots,d \} \big\}$. On the other hand, in similar prior works, the sample complexity requirement has been shown to be either linear or super-linear in $\bar{n}$~\cite{tomioka2011statistical, mu2014square, rauhut2017}. We also employ synthetic data experiments to demonstrate the efficacy of the proposed computational procedure. Finally, we conduct real-data experiments on a collection of fMRI images pertaining to attention deficit hyperactivity disorder~\cite{milham2012adhd}, and we show that the imposition of multiple structures on $\B$ allows for efficient neuroimaging analysis in the low sample size regime.

\subsection{Notation}\label{subsec:notation}
Bold upper-case letters ($\mathbf{Z}$), upper-case letters ($Z$), bold lower-case letters ($\mathbf{z}$), lower-case letters ($z$), and underlined letters ($\barbelow{z}$) are used to denote tensors, matrices, vectors, scalars, and tuples, respectively. For any tuple $\barbelow{z}$ and scalar $\alpha$, we use $\alpha \barbelow{z}$ to denote the tuple obtained by multiplying each entry of $\barbelow{z}$ by $\alpha$. For any scalar $q\in\mathbb{Z}_{+}$, we use $[[q]]$ as a shorthand for $\{1,2,\dots,q\}$. Given any vector $\mathbf{u}\in \mathbb{R}^n$, $\|\mathbf{u}\|_0$ and $\|\mathbf{u}\|_2$ denote the $\ell_0$ and $\ell_2$ norms of vector $\mathbf{u}$, respectively. Given two vectors $\mathbf{u}\in \mathbb{R}^n$ and $\mathbf{v}\in \mathbb{R}^n$ of same dimension, $\mathbf{u} \circ \mathbf{v}$ denotes their outer product. Given any matrix $\U$, the $i$-th column is denoted by $\U(:, i)$, the spectral norm is denoted by $\| \U \|_{2}$, while $\max \limits_{i} \|\U(:,i)\|_2$ is denoted by $\|U\|_{1,2}$. Given any two matrices $\U_1$ and $\U_2$, $\U_1 \otimes \U_2$ denotes the Kronecker product. Given any tensor $\Z$, its $(i_1,i_2,\ldots,i_d)$-th entry is given by $\Z(i_1, i_2, \ldots, i_d)$, the Frobenius norm $\|\Z\|_F$ is given by $\sqrt{\sum \limits_{i_1, i_2, \ldots, i_d} \Z (i_1, i_2, \ldots, i_d)^2}$, the $\ell_1$ norm $\|\Z\|_1$ is given by $\sum \limits_{i_1, i_2, \ldots, i_d} | \Z (i_1, i_2, \ldots, i_d)|$, and the mode$\mhyphen i$ matricization $\Z_{(i)}$ is the matrix obtained from column-arrangement of the mode$\mhyphen i$ fibers of $\Z$. The conjugate transpose of a linear map $\A: \mathbb{R} ^ {n_1 \times n_2 \times \dots \times n_d } \rightarrow \mathbb{R} ^ {m}$ is denoted by $\A^{*}: \mathbb{R} ^ {m}  \rightarrow \mathbb{R} ^ {n_1 \times n_2 \times \dots \times n_d }$. Following the tensor notation in \cite{kolda2009tensor}, for matrices $\widetilde{\U}_i \in \mathbb{R} ^ {n_i \times r_i}$, $i\in [[d]]$, and tensor $\Score \in \mathbb{R} ^ {r_1 \times r_2 \times \dots \times r_d }$, we define $\Score \times_1 \widetilde{\U}_1 \times_2 \widetilde{\U}_2 \cdots \times_d \widetilde{\U}_d$ as ${\sum \limits_{i_1, i_2, \ldots, i_d} \Score (i_1, i_2, \ldots, i_d) \; \widetilde{\U}_1(:, i_1) \circ \widetilde{\U}_2(:, i_2) \circ \cdots \circ \widetilde{\U}_d(:, i_d)}$. Finally, $\mathbb{I}_q$ refers to an identity matrix of size $q$, where $q\in\mathbb{Z}_{+}$.

\subsection{Organization}
The rest of this paper is organized as follows. In Sec.~\ref{sec:PF}, we describe the regression model, which includes a formal definition of sparse, low Tucker-rank tensors, and then we present a non-convex formulation of the problem for estimating the regression tensor. In Sec.~\ref{sec:ProposedSolution}, we propose a method for solving the posed non-convex problem, and in Sec.~\ref{sec:Convergence}, we provide mathematical guarantees for the proposed method, based on a certain restricted isometry property of the predictor tensors. In Sec.~\ref{sec:Analysis}, we evaluate the posed property for sub-Gaussian predictors and provide sample complexity bounds. In Sec.~\ref{sec:Experiments}, we report results of extensive numerical experiments on both synthetic and real data, while concluding remarks are presented in Sec.~\ref{sec:Conclusion}. Finally, in the interest of space, we present some of the proofs not central to understanding of the implications of this work in the appendix.

\section{Problem formulation}\label{sec:PF}
For ease of notation, let us define $\mathcal{W}:= \mathbb{R} ^ {n_1 \times n_2 \times \dots \times n_d }$, $\mathcal{Y}:= \mathbb{R} ^ {m}$, and let us denote the collection of tensors $\{ \X_i \}_{i=1}^{m}$ in \eqref{eq:System1} by a linear map/measurement operator $\A: \mathcal{W} \rightarrow \mathcal{Y}$ such that \eqref{eq:System1} can equivalently be expressed as
\begin{align}\label{eq:Measurement}
\y = \A( \W ) + \mathbf{\noise} \text{,}
\end{align}
where $\y = [ y_1, y_2, \ldots, y_m ]$, and $\mathbf{\noise} = [ \eta_1, \eta_2, \ldots, \eta_m ]$. In this work, we impose that the parameter tensor $\W \in \mathbb{R} ^ {n_1 \times n_2 \times \dots \times n_d }$ is structured in the sense that it is $\rank$-rank and $\sparsity$-sparse\revise{, simultaneously. We formally define the notion of an $\rank$-rank and $\sparsity$-sparse tensor as follows.}

\begin{definition}[$\rank$-rank and $\sparsity$-sparse tensor]\label{def:LowRankSparse}
Given a rank tuple $\rank := (r_{1}, r_{2}, \dots, r_{d})$ and a sparsity tuple $\sparsity := (s_{1}, s_{2}, \dots, s_{d})$, a tensor $\Z \in \mathbb{R} ^ {n_1 \times n_2 \times \dots \times n_d }$ is said to be \revise{both} $\rank$-rank and $\sparsity$-sparse if $\Z$ can be expressed as
\begin{align}\label{eq:Model}
\Z = \Score \times_1 \U_1 \times_2 \U_2 \cdots \times_d \U_d \text{,}
\end{align}
where $\Score \in \mathbb{R} ^ {r_1 \times r_2 \times \dots \times r_d }$ and $\U_i \in \mathbb{R} ^ {n_i \times r_i}$, with $\| \U_i(:,j) \|_0\leq s_i$, $\forall i\in [[d]]$, $j\in [[r_i]]$. Notice that, trivially, $r_i \leq n_i$ and $s_i \leq n_i$.
\end{definition}

Recall from \cite{kolda2009tensor} that \eqref{eq:Model} is expressing $\Z$ in terms of a Tucker decomposition, in which $\Score$ is termed the core tensor and the $\U_i$'s are referred to as factor matrices, with additional sparsity constraints on the factor matrices. It can also be seen from \eqref{eq:Model} that for the special case when $s_i=n_i$, the mode$\mhyphen i$ matricization of $\Z$ has rank $r_i$: rank($\Z_{(i)}$) $= r_i$; i.e., the $\rank \mhyphen$rank of $\Z$ is simply the Tucker rank of $\Z$. Further, note that we are defining sparsity of $\Z$ in terms of sparsity of the columns of the factor matrices $\{ \U_i(:,j)\}$, $ i\in[[d]],\; j\in r_i$, that are generating the tensor. This notion of sparsity is different from the conventional notion of sparsity, where sparsity is defined as the number of non-zero entries for the data structure under consideration, i.e., tensor $\Z$ in this case. In contrast, the \revise{notion of sparsity in Definition \ref{def:LowRankSparse}} not only induces sparsity on $\Z$, but it also dramatically reduces the number of free parameters in $\Z \in \mathbb{R} ^ {n_1 \times n_2 \times \dots \times n_d }$ from $n:=\prod \nolimits_{i} n_i$ to the order of $ \prod \nolimits_{i} r_{i} + \sum \nolimits_{i} r_{i} s_{i} \log n_i$, which can be significantly smaller than $n$ for $r_i \ll n_i$ and $s_i \ll n_i$. \revise{(Note that the $\log n_i$ factor arises from the need to encode the locations of the $s_i$ non-zero entries in a given column of $\U_i$.)} This reduction in degrees of freedom allows us to learn the tensor regression model in \eqref{eq:Measurement} with lower sample complexity, as we show later.

Since we are requiring the unknown tensor $\W$ to be $\rank$-rank and $\sparsity$-sparse in our regression model~\eqref{eq:Measurement}, we formally define a set of such tensors as follows:
\begin{align}
\label{eq:model_main}
\Classzero = & \{ \Score \times_1 \U_1 \times_2 \U_2 \times_3 \cdots \times_d \U_d:
\Score \in \mathbb{R} ^ {r_1 \times r_2 \times \dots \times r_d }, \text{ and }
\nonumber\\
& \U_i \in \mathbb{R} ^ {n_i \times r_i}, \| \U_i(:,j) \|_0\leq s_i, i\in [[d]], j\in [[r_i]]  \} \text{.}
\end{align}

\noindent Using the definition of constraint set $\Classzero$, and given a known linear map $\A$, we can pose the following constrained optimization problem for recovery of $\W$ from noisy observations $\y$:
\begin{align}\label{eq:Objective}
\hat{\B} = \arg \min_{\Z\in \Classzero} \frac{1}{2} \; \|\y - \A(\Z)\|_2^2 \text{.}
\end{align}
We can see that the optimization problem posed in \eqref{eq:Objective} is non-convex because of non-convexity of the constraint set $\Classzero$. In contrast, most of the prior works in tensor parameter estimation focus on solving convex relaxations of the tensor recovery problem for various notions of low-dimensional tensor structures~\cite{liu2009tensor, gandy2011tensor, tomioka2011statistical, mu2014square}, hence benefiting from rich literature on theory and algorithms for convex optimization. But the issue with convex relaxation-based solutions is that \emph{convex relaxations can be suboptimal in terms of number of measurements required to solve the problem} \cite{mu2014square}. While posing and solving the tensor recovery problem in a non-convex form tends to circumvent this issue, it brings about difficulties in terms of theoretically characterizing behavior of the associated recovery algorithm. In the next section, we present our proposed method for solving \eqref{eq:Objective}, while theoretical characterization of the proposed approach follows in Sec.~\ref{sec:Convergence} and Sec.~\ref{sec:Analysis}.

\section{Estimation of $\rank \mhyphen$rank and $\sparsity \mhyphen$sparse Regression Tensors}\label{sec:ProposedSolution}
In this section, we present a method for estimation of the structured parameter tensor $\W$ in the regression model \eqref{eq:Measurement}, given the linear map $\A$, response vector $\y$, and the assumption that $\W$ is $\rank \mhyphen$rank and $\sparsity \mhyphen$sparse. Our method is inspired by the various projected gradient descent-based methods in the literature, where such methods have been employed for recovery of sparse vectors \cite{blumensath2009iterative}, low-rank matrices \cite{jain2010guaranteed}, and more recently, low-rank tensors \cite{rauhut2017, yu2016learning}. The method, termed \emph{tensor projected gradient descent} (TPGD), is summarized in Algorithm \ref{algo:TPG}. The TPGD method consists of two steps. First we perform gradient descent iteration over the objective function in~\eqref{eq:Objective} (Step~4, Algorithm~\ref{algo:TPG}), and then we project the iterate onto set $\Classzero$, which is the set of $\rank$-rank and $\sparsity$-sparse tensors (Step~5, Algorithm~\ref{algo:TPG}). The projection operator, $\Hsr: \mathbb{R} ^ {n_1 \times n_2 \times \dots \times n_d } \rightarrow \mathbb{R} ^ {n_1 \times n_2 \times \dots \times n_d }$, in Step~5 of Algorithm \ref{algo:TPG} is defined as:
\begin{align}
\Hsr ( \Wtilde ) := \arg\min\limits_{\What\in \Classzero} \| \Wtilde - \What \|_F^2.
\label{eq:Hsr}
\end{align}

\begin{algorithm}[t]
	\algsetup{indent=1em}
	\begin{algorithmic}[1]
		\STATE \textbf{Input:} Linear map $\A$, response vector $\y$, step size $\mu$, sparsity tuple $\sparsity$, rank tuple $\rank$
		\STATE \textbf{Initialize:} Tensor $\W^0$ and $k\leftarrow 0$
		\WHILE {Stopping criterion}
		\STATE $\Wtildek \leftarrow \Wk - \mu \A^{*}(\A(\Wk) - \y)$
		\STATE $\Wkplus \leftarrow \Hsr(\Wtildek)$
		\STATE $k\leftarrow k+1$
		\ENDWHILE
		\RETURN Tensor $\Walgo = \Wk$
		\caption{Tensor Projected Gradient Descent (TPGD)}
		\label{algo:TPG}
	\end{algorithmic}
\end{algorithm}

In general, computation of the best low-rank approximation of a given tensor is considered to be an NP-hard problem~\cite{hillar2013most, rauhut2017}. Despite that, several algorithms have been proposed in the literature for computing low-rank tensor approximations corresponding to various notions of tensor decompositions~\cite{kolda2009tensor, allen2012sparse, grasedyck2013literature, 2017provable}.
Although these approximation algorithms do not come with mathematical guarantees regarding the accuracy of tensor approximation, they have been employed successfully in practice for tensor approximation within various methods for estimating tensor-structured parameters in regression models~\cite{yu2016learning,rauhut2017}. The mathematical guarantees for these parameter estimation methods \emph{assume} the goodness of the tensor approximation step, since the corresponding approximation algorithms are not guaranteed to compute the best approximation.

In a similar vein, in the mathematical guarantees for Algorithm~\ref{algo:TPG} (Sec.~\ref{sec:Convergence}), we \emph{assume} that the best low-rank and sparse approximation (projection step in Step~5, Algorithm~\ref{algo:TPG}) can be exactly computed. However, in our numerical simulations (Sec.~\ref{sec:Experiments}), we employ Algorithm~\ref{algo:TPGprojection} for computation of the projection step, where Algorithm~\ref{algo:TPGprojection} is essentially the Sparse Higher-Order SVD method~\cite{allen2012sparse}. Moreover, within Algorithm~\ref{algo:TPGprojection}, we employ \cite{hein2010inverse} for computation of the factor matrices $\{ \bar{U}_j \}_{j=1}^{d}$ (Step~3, Algorithm~\ref{algo:TPGprojection}). Later, in Sec.~\ref{sec:Experiments}, our numerical simulations show that Algorithm~\ref{algo:TPGprojection} can indeed be effectively employed with Algorithm~\ref{algo:TPG} to efficiently learn the regression model in \eqref{eq:Measurement} under certain conditions, despite the lack of mathematical guarantees for Algorithm~\ref{algo:TPGprojection}.

\begin{algorithm}[h]
	\algsetup{indent=1em}
	\begin{algorithmic}[1]
		\STATE \textbf{Input:} Tensor $\Wtilde$, sparsity tuple $\sparsity$, rank tuple $\rank$
		\FOR {$j = 1, \ldots, d$}
		\STATE $ \bar{U}_j \leftarrow \text{ First } r_j, s_j \mhyphen \text{sparse principal components of } \Wtilde_{(j)}$
		\ENDFOR
		\STATE $\bar{\Score} \leftarrow \Wtilde \times_1 \bar{U}_1 \times_2 \bar{U}_2 \times_3 \cdots \times_d \bar{U}_d $
		\RETURN Tensor $\bar{\W} = \bar{\Score} \times_1 \bar{U}_1 \times_2 \bar{U}_2 \times_3 \cdots \times_d \bar{U}_d $
		\caption{Sparse Higher-Order SVD}
		\label{algo:TPGprojection}
	\end{algorithmic}
\end{algorithm}

\section{Convergence Analysis of Tensor Projected Gradient Descent}\label{sec:Convergence}
In this section we provide theoretical guarantees for TPGD (Algorithm~\ref{algo:TPG}), which, as explained earlier, is a projected gradient method to solve~\eqref{eq:Objective}. Variants of the projected gradient method have been analyzed for recovery of sparse vectors \cite{blumensath2009iterative}, low-rank matrices \cite{jain2010guaranteed}, and low-rank tensors \cite{yu2016learning, rauhut2017, chen2016non} under the assumption that the linear map/measurement operator satisfies some variant of the restricted isometry property (RIP) \cite{candes2008restricted}. Since different tensor decompositions induce different notions of tensor rank~\cite{rauhut2017, hao2018sparse}, and different regression models lead to different measurement operators \cite{rauhut2017, yu2016learning}, various notions of RIP have also been posed for various tensor decompositions and regression models. Before we present the notion of RIP assumed on the linear map in this work, let us define a set of $\rank$-rank and $\sparsity$-sparse tensors, with additional constraints on ($i$) the $\ell_1\mhyphen$norm of the associated core tensor and ($ii$) the $\ell_2\mhyphen$norm of columns of the associated factor matrices:
\begin{align}
\label{eq:model_main_2}
\Class = & \{ \Score \times_1 \U_1 \times_2 \U_2 \times_3 \cdots \times_d \U_d:
\Score \in \mathbb{R} ^ {r_1 \times r_2 \times \dots \times r_d }, \|\Score\|_1\leq \corebound, \text{ and }
\nonumber\\
& \U_i \in \mathbb{R} ^ {n_i \times r_i}, \| \U_i(:,j) \|_0\leq s_i, \| \U_i(:,j) \|_2 \leq 1, i\in [[d]], j\in [[r_i]]  \} \text{.}
\end{align}
From the $\ell_1\mhyphen$norm constraint on $\Score$ and $\ell_2 \mhyphen$norm constraint on $\U_i(:,j)$, where $i \in [[d]]$ and $j \in [[r_i]]$, it follows that $\|Z\|_F\leq \tau$ for any $Z \in \Class$. \revise{These norm constraints in \eqref{eq:model_main_2} allow us to bound the covering number of the set $\Class$, which enables us to obtain a sample complexity bound for tensor recovery, as follows in the next section. Specifically, in order to derive a bound on the covering number of $\Class$ in the next section, our mathematical analysis requires bounds on $\|\Score\|_1$,  $\|Z\|_F$, and $\| \U_i(:,j) \|_2$ for $i\in [[d]]$, $j\in [[r_i]]$. Since the $\ell_1\mhyphen$norm constraint on $\Score$ and the $\ell_2 \mhyphen$norm constraint on $\U_i(:,j)$ result in $\|Z\|_F \leq \tau$, the constraints in \eqref{eq:model_main_2} suffice to evaluate a bound on the covering number of $\Class$.}

For the recovery of $\rank$-rank and $\sparsity$-sparse tensors considered in this work, we consider the following notion of RIP on the linear map $\A$.
\begin{definition}[$(\rank, \sparsity, \corebound, \deltarip)\mhyphen$Restricted Isometry Property]
The restricted isometry constant $\deltarip\in(0, 1)$ of a linear map $\A: \mathbb{R} ^ {n_1 \times n_2 \times \dots \times n_d } \rightarrow \mathbb{R} ^ {m}$ acting on tensors of order $d$ is the smallest quantity such that
\begin{align}
(1-\deltarip) \| \Z \|_F^2 \leq \| \A (\Z) \|_2^2 \leq (1+\deltarip) \| \Z \|_F^2
\label{eq:RIP}
\end{align}
for all tensors $\Z \in \Class $.
\label{def:RIP}
\end{definition}
 In the following we provide our first main theoretical result that characterizes the convergence behavior of TPGD under the assumption of an exact projection step (Step~5, Algorithm~\ref{algo:TPG}).

\begin{theorem}\label{th:main_result}[Convergence of TPGD]
Let $\y=\A( \W )+\noise$, and let $\W^{0} \in\Class$ be the tensor initialization in Algorithm~\ref{algo:TPG}. For some fixed $\gamma \in(0, 1)$, \revise{suppose $\A: \mathbb{R} ^ {n_1 \times n_2 \times \dots \times n_d } \rightarrow \mathbb{R} ^ {m}$ satisfies RIP in Definition \ref{def:RIP} with $\deltariptwo < \frac{\gamma}{4+\gamma}$. Then, fixing the step size $\mu = \frac{1}{1+\deltariptwo}$ and defining $b := \frac{1+3 \deltariptwo}{1 - \deltariptwo}$, the estimation error in TPGD algorithm's (Algorithm~\ref{algo:TPG}) iterate, $\mathbf{B}^k$, after $k$ iterations is given by:
		\begin{align*}
		\|\mathbf{B}^k - \mathbf{B}\|_F^2\leq \frac{2 \gamma^k}{1-\deltariptwo}\left\| \y - \A(\mathbf{B}^0)\right\|_2^2+\frac{2\|\noise\|_2^2}{1-\deltariptwo}\Big(1+\frac{b}{1-\gamma}\Big).
		\end{align*}
}
\end{theorem}

\subsection{Discussion of Theorem \ref{th:main_result}}
\revise{Let $c_0 := \frac{2\|\noise\|_2^2}{1 - \deltariptwo}\Big(1+\frac{b}{1-\gamma}\Big)$. Next, define the closed ball $\mathcal{B}(c_0,\W)$ with center at $\W$ and radius $c_0$ as the set of all the tensors $\Z \in \mathbb{R} ^ {n_1 \times n_2 \times \dots \times n_d }$ such that $\| \Z - \W \|_F^2 \leq c_0$. Theorem~\ref{th:main_result} shows that starting from an initial estimate $\mathbf{B}^0$, the solution of TPGD converges linearly to the set $\mathcal{B}(c_0,\W)$ at the rate of $\gamma^k$. Additionally, Theorem~\ref{th:main_result} also characterizes the impact of noise power $\|\noise\|_2^2$ and RIP constant $\deltariptwo$ on the convergence behavior of the TPGD algorithm. First, the radius of ball $\mathcal{B}(c_0, \W)$ scales linearly with the noise power $\|\noise\|_2^2$. Thus, the more the noise power, the less accurate may the solution of TPGD be and vice versa. Second, Theorem~\ref{th:main_result} shows that the smaller the RIP constant $\deltariptwo$, the smaller the radius of ball $\mathcal{B}(c_0, \W)$. Thus, the larger the value of $\deltariptwo$, the less accurate may the solution of TPGD be and vice versa. We conclude by noting that although the mathematical guarantees in Theorem~\ref{th:main_result} depend on the $(\rank, \sparsity, \corebound, \deltarip)\mhyphen$RIP property in Definition~\ref{def:RIP}, we evaluate this property for a known family of linear maps in the next section.}

\subsection{Remarks on Proof of Theorem \ref{th:main_result}}
A key step in proving Theorem~\ref{th:main_result} is to show that any linear combination of two $\rank$-rank and $\sparsity$-sparse tensors has rank at most $2\rank$ and sparsity $\sparsity$. We formally describe this in terms of a lemma that appears in analysis of any step that involves linear combination of $\rank$-rank and $\sparsity$-sparse tensors.
\begin{lemma}\label{lemma:add_tensors}
Let $\Z_a \in \mathbb{R}^{n_1 \times n_2 \times \cdots \times n_d}$ and $\Z_b\in \mathbb{R}^{n_1 \times n_2 \times \cdots \times n_d}$ be members of the set $\Class$, where $\rank := (r_1, r_2, \ldots, r_d)$, $\sparsity := (s_1, s_2, \ldots, s_d)$, and $\corebound \in \mathbb{R}^{+}$. Define $\Z_c = \gamma_a \Z_a + \gamma_b \Z_c$, where $\gamma_a, \gamma_b \in \mathbb{R}$. Then, $\Z_c$ is a member of the set $\mathcal{G}_{2\rank,\sparsity, \kappa }$, where $\kappa = (|\gamma_a| + |\gamma_b|) \corebound$.
\end{lemma}
The proof of this lemma is provided in Appendix~\ref{app:proof_add_tensors}, and the proof of Theorem~\ref{th:main_result} follows in Appendix~\ref{app:main_result}.

\section{Evaluating the Restricted Isometry Property for Sub-Gaussian Linear Maps}\label{sec:Analysis}
In the previous section, we provided theoretical guarantees for recovery of the parameter tensor $\B$ using the TPGD method, based on assumption of the Restricted Isometry Property (Definition~\ref{def:RIP}). In this section, we provide examples of linear maps that satisfy this property. Specifically, we consider linear maps in \eqref{eq:Measurement}, $\A$, that denote the collection of tensors in \eqref{eq:System1}, $\{ \X_i \}_{i=1}^{m}$, such that the entries of \revise{each $\X_i$} are independently drawn from zero-mean, unit-variance sub-Gaussian distributions. We term such linear maps as sub-Gaussian linear maps. Before we evaluate the condition in Definition~\ref{def:RIP} for these maps, let us recall the definition of a sub-Gaussian random variable.
\begin{definition}
A zero-mean random variable $\mathcal{Z}$ is said to follow a sub-Gaussian distribution $subG(\alpha)$ if there exists a sub-Gaussian parameter $\alpha>0$ such that $\mathbb{E}[\exp (\lambda \mathcal{Z})] \leq \exp \Big(\frac{\alpha^2 \lambda^2}{2}\Big)$ for all $\lambda\in\mathbb{R}$.
\end{definition}
\noindent In words, a $subG(\alpha)$ random variable is one whose moment generating function is dominated by that of a Gaussian random variable. Some common examples of sub-Gaussian random variables include:
\begin{itemize}
\item $\mathcal{Z} \sim$ $\mathcal{N}(0,\alpha^2) \ \Rightarrow \ \mathcal{Z} \sim subG(\alpha)$.
\item $\mathcal{Z} \sim \text{unif}(-\alpha,\alpha) \ \Rightarrow \ \mathcal{Z} \sim subG(\alpha)$.
\item $|\mathcal{Z}| \leq \alpha, \mathbb{E}[\mathcal{Z}]=0 \ \Rightarrow \ \mathcal{Z} \sim subG(\alpha)$.
\item $\mathcal{Z} \sim
\begin{cases}
		  \alpha,& \text{ with prob. } \frac{1}{2},\\
		 -\alpha,& \text{ with prob. } \frac{1}{2},\\
\end{cases} \ \Rightarrow \ \mathcal{Z} \sim subG(\alpha)$.
\end{itemize}
We now evaluate the Restricted Isometry Property (Definition~\ref{def:RIP}) for sub-Gaussian linear maps. An outline of the proof of the following result is provided in Sec.~\ref{subsec:ProofTheorem}, while its detailed proof is part of Appendix~\ref{app:proof_subGaussian}.

\begin{theorem}
Let the entries of $\{ \X_i \}_{i=1}^{m}$ be independently drawn from zero-mean, $\frac{1}{m}$-variance $subG(\alpha)$ distributions. Define $\bar{n}:=\max \{n_i: i\in [[d]] \}$. Then, for any $\delta, \varepsilon \in (0,1)$, the linear map $\A$ satisfies $\deltarip \leq \delta$ with probability at least $1 - \varepsilon$ as long as
\begin{align*}
m \geq  \delta^{-2} \, \max \Bigg\{
K_1 \, \corebound^2 \, \bigg( \prod \limits_{i=1}^{d} r_i + \sum \limits_{i=1}^{d} s_i r_i \bigg) \Big(\log(3 \bar{n} d)\Big)^2,
\; K_2 \, \log(\varepsilon^{-1})
\Bigg\} \text{,}
\end{align*}
where the constants $K_1$, $K_2 > 0$ depend on $\corebound$ and $\alpha$.
\label{th:subGaussian}
\end{theorem}

\subsection{Discussion}\label{subsec:Discussion}
We compare the result in Theorem~\ref{th:subGaussian} with sample complexity bounds in the literature for estimation of the parameter tensor $\B$ in \eqref{eq:Measurement}. Theoretically, we can pose the estimation problem as ($i$) low Tucker-rank recovery problem~\cite{rauhut2017}, or ($ii$) sparse recovery problem~\cite{rish2014sparse}. Thus, in this section, we first compare the sample complexity bound in Theorem~\ref{th:subGaussian} with complexity bounds from low rank recovery and sparse recovery literature. For ease of comparison, define $\bar{r}:= \max \{r_1,r_2,\ldots,r_d\}$ and $\bar{s}:= \max \{s_1,s_2,\ldots,s_d\}$. With these definitions, the sample requirement in Theorem~\ref{th:subGaussian} can be written as $\mathcal{O} \Big( \big(\bar{r}^d + \bar{s} \, \bar{r} \, d \big) \big(\log (3\, \bar{n} \, d) \big)^2 \Big)$. We now compare this result with complexity bounds in prior works.

\subsubsection{Low Tucker-rank recovery}\label{subsec:lowTucker}
Among the many works that study the problem of estimating $\B$ under the imposition of low Tucker rank on $\B$~\cite{gandy2011tensor, tomioka2011statistical, mu2014square, rauhut2017}, the most tight sample complexity bound has been shown to be $\mathcal{O} \big( (\bar{r}^d + \bar{n} \, \bar{r} \,d) \log (d) \big)$~\cite{rauhut2017}. If we apply this complexity bound for estimating the parameter tensor $\B$ in \eqref{eq:Measurement}, the sample complexity requirement scales linearly with $\bar{n}$. In contrast, since we consider sparsity on columns of the factor matrices within Tucker decomposition of $\B$, our sample complexity bound has a linear dependence on $\bar{s}$ and only a polylogarithmic dependence on $\bar{n}$, where $\bar{s} \ll \bar{n}$.

\subsubsection{Sparse recovery}\label{subsec:sparsity}
The regression model in \eqref{eq:Measurement}, or equivalently the model in \eqref{eq:System1}, can be vectorized such that the model can be expressed as $y_i = \langle \text{vec}(\X_i) , \text{vec}(\B) \rangle + \eta_i$, $i\in [[m]]$, and the problem of recovering $\B$ can be posed as a sparse recovery problem. It has been shown that if the entries of $\text{vec}(\X_i)$, $i\in[[m]]$, draw values from a Gaussian distribution, $\text{vec}(\B)$ can be recovered using $\mathcal{O}(k \log(\bar{n}^{d}/k))$ samples~\cite{ba2010lower}, where $k$ is the number of non-zero entries in $\text{vec}(\B)$. The number of non-zero entries in $\text{vec}(\B)$ are upper bounded by $(\bar{s} \, \bar{r} )^d$, which leads to the worst-case sample complexity requirement of $\mathcal{O}( d \, (\bar{s} \, \bar{r})^d \log(\bar{n} / \bar{s} \, \bar{r}))$. Thus, the sparse signal recovery literature poses a worst-case sample complexity requirement that has linear dependence on $d \, (\bar{s} \,\bar{r} )^d$. In contrast, since we consider the multi-dimensional structure within $\B$, our sample complexity requirement has only linear dependence on $\bar{r}^d + \bar{s} \, \bar{r} \, d$.

Finally, note that the number of free parameters in the parameter tensor $\B$ are on the order of $ \prod \nolimits_{i} r_{i} + \sum \nolimits_{i} r_{i} s_{i} \log n_i$, \revise{where the $\log n_i$ factor encodes for the $s_i$ non-zero entries in each of the $r_i$ columns of the $i$-th factor matrix of the tensor $\B$. More compactly, this number of free parameters can be expressed as $\bar{r}^d + \bar{s} \, \bar{r} \, d \log \bar{n}$.} Thus, the posed sample complexity requirement of $\mathcal{O} \Big( \big(\bar{r}^d + \bar{s} \, \bar{r} \, d \big) \big(\log (3\, \bar{n} \, d) \big)^2 \Big)$ in Theorem~\ref{th:subGaussian} is order-optimal up to a polylogarithmic factor.

\subsection{Outline of the Proof}\label{subsec:ProofTheorem}
The general idea of the proof of Theorem~\ref{th:subGaussian} is similar to that of \cite[Theorem~4.2]{jain2010guaranteed}, \cite[Theorem~2.3]{candes2011tight}, \cite[Theorem~4.1]{krahmer2014suprema}, and \cite[Theorem~2]{rauhut2017}, where the main analytic challenge is to analyze the complexity of the set that is hypothesized to contain the regression parameters. In this work, the challenge translates into characterizing the complexity of the set $\Class$, for which we employ the notion of $\epsilon$-nets and covering numbers, defined as follows.

\begin{definition}[$\epsilon$-nets and covering numbers]
Let $( \mathbb{V} , h )$ be a metric space, and let $\mathbb{T} \subset \mathbb{V}$. The set $X \subset \mathbb{T}$ is called an $\epsilon$-net of $\mathbb{T}$ with respect to the metric $h$ if for any $T_i \in \mathbb{T}$, $\exists X_i \in X$ such that $h(X_i, T_i) \leq \epsilon$. The minimum cardinality of an $\epsilon$-net of $\mathbb{T}$ (with respect to the metric $h$) is called the covering number of $\mathbb{T}$ with respect to the metric $h$ and is denoted by $\covering(\mathbb{T}, h, \epsilon)$ in this paper.
\label{def:covering_number}
\end{definition}

\noindent Next, we provide an outline to the proof of Theorem~\ref{th:subGaussian}. In the first step, we provide an upper bound on the covering number of $\Class$ with respect to the Frobenius norm, which forms our main contribution. In the second step, we employ a deviation bound from prior works~\cite{rauhut2017,krahmer2014suprema} to complete the proof of this theorem. A formal proof of Theorem~\ref{th:subGaussian} follows in Sec.~\ref{app:proof_subGaussian}.

\subsubsection{Bound on covering number of $\Class$}
\label{subsubsec:Covering}
\noindent The following lemma provides a bound on the covering number of $\Class$ with respect to the Frobenius norm.

\begin{lemma}
For tuples $\rank := (r_{1}, r_{2}, \dots, r_{d})$, $\sparsity := (s_{1}, s_{2}, \dots, s_{d})$, and for any $\corebound>0$, the covering number of
\begin{align*}
\Class = & \{ \Score \times_1 \U_1 \times_2 \U_2 \times_3 \cdots \times_d \U_d:
\Score \in \mathbb{R} ^ {r_1 \times r_2 \times \dots \times r_d }, \|\Score\|_1\leq \corebound, \text{ and }
\\
& \U_i \in \mathbb{R} ^ {n_i \times r_i}, \|U(:,j)\|_2 \leq 1, \| \U_i(:,j) \|_0\leq s_i, i\in [[d]], j\in [[r_i]]  \}
\end{align*}
with respect to the metric $\dist_{\mathcal{G}}$ satisfies
\begin{align*}
\covering(\Class, \dist_{\mathcal{G}}, \epsilon) \leq
\Big( \frac{3 \corebound (d+1)}{\epsilon} \Big)^{\prod \limits_{i=1}^{d} r_i} \Big( \frac{3  \bar{n}  \corebound (d+1)}{\epsilon} \Big)^{\sum \limits_{i=1}^{d} s_i r_i}, \epsilon \in (0,1)\text{,}
\end{align*}
where $\bar{n}:=\max \{n_i: i\in [[m]] \}$ and $\dist_{\mathcal{G}} (\mathbf{G}^{(1)}, \mathbf{G}^{(2)}) = \| \mathbf{G}^{(1)} - \mathbf{G}^{(2)}\|_F$ for any $\mathbf{G}^{(1)}, \mathbf{G}^{(2)} \in \Class$.
\label{lemma:low_rank_sparse_covering}
\end{lemma}

\noindent Let us provide an outline to the proof of Lemma~\ref{lemma:low_rank_sparse_covering}, while a formal proof is provided in Sec.~\ref{app:low_rank_sparse_covering}. Define Cartesian product of metric spaces $(\mathcal{D}_{\Score}, \dist_{\Score})$, $(\mathcal{D}_{U_1}, \dist_{U_1})$, $(\mathcal{D}_{U_2}, \dist_{U_2})$, $\ldots$ , $(\mathcal{D}_{U_d}, \dist_{U_d})$, that is
\begin{align}
\mathcal{D}_{P} := \mathcal{D}_{\Score} \times \mathcal{D}_{U_1} \times \mathcal{D}_{U_2} \times \cdots \times \mathcal{D}_{U_d} \text{,}
\label{eq:product_set}
\end{align}
where $\mathcal{D}_{\Score} := \{ \Score \in \mathbb{R}^{r_1 \times r_2 \times \cdots \times r_d}: \|\Score\|_1 \leq \tau \}$, $\dist_{\Score}(\Score^{(1)},\Score^{(2)}) := \frac{1}{\corebound} \| \Score^{(1)} - \Score^{(2)} \|_1$ for any $\Score^{(1)}, \Score^{(2)} \in \mathcal{D}_{\Score}$, $\mathcal{D}_{U_i} := \{ U\in \mathbb{R}^{n_i \times r_i}: \|U(:,j)\|_2 \leq 1, \|U(:,j)\|_0 \leq s_i, j\in [[r_i]]\}$, and $\dist_{U_i} (U_i^{(1)}, U_i^{(2)}) = \| U_i^{(1)} - U_i^{(2)} \|_{1,2}$ for any  $U_i^{(1)}, U_i^{(2)} \in \mathcal{D}_{U_i}$, for all $i\in [[d]]$.
First, we need to compute an upper bound on the covering number of $\mathcal{D}_{P}$ with respect to the metric $\dist_{P}$ defined as
\begin{align}
\dist_{P}(P^{(1)}, P^{(2)}) = \max \big\{  \max \limits_{i\in [[d]]} \{  \dist_{U_i} (U_i^{(1)}, U_i^{(2)}) \}, \dist_{\Score}(\Score^{(1)},\Score^{(2)}) \big\} \text{,}
\label{eq:product_metric}
\end{align}
where $P^{(1)}, P^{(2)}\in \mathcal{D}_{P}$, $\Score^{(1)}, \Score^{(2)} \in \mathcal{D}_{\Score}$, and $U_i^{(1)}, U_i^{(2)} \in \mathcal{D}_{U_i}$, $i\in[[d]]$. Specifically, using Lemma~\ref{lemma:covering_cartesian}, a bound on $\covering(\mathcal{D}_{P}, \dist_{P}, \epsilon)$ can be obtained as
\begin{align}
\covering(\mathcal{D}_{P}, \dist_{P}, \epsilon)  \leq
\covering(\mathcal{D}_{\Score}, \dist_{\Score}, \epsilon) \;
\prod \limits_{i=1}^{d} \covering(\mathcal{D}_{U_i}, \dist_{U_i}, \epsilon) \text{.}
\label{bound:Cartesian}
\end{align}
Thus, to compute an upper bound on $\covering(\mathcal{D}_{P}, \dist_{P}, \epsilon)$, we need upper bounds on $\covering(\mathcal{D}_{\Score}, \dist_{\Score}, \epsilon)$ and $\covering(\mathcal{D}_{U_i}, \dist_{U_i}, \epsilon)$, respectively. To obtain a bound on $\covering(\mathcal{D}_{\Score}, \dist_{\Score}, \epsilon)$, we employ the following lemma, which is proved in Appendix~\ref{app:CoveringNumber_HS}.

\begin{lemma}
Define $\mathcal{D}_{\Score} := \{ \Score \in \mathbb{R}^{r_1 \times r_2 \times \cdots \times r_d} : \|\Score \|_1 \leq \corebound \}$ with distance measure $\normed_1$. Then the covering number of $\mathcal{D}_{\Score}$ (with respect to the norm $\normed_1$) satisfies the bound
\begin{align*}
\covering (\mathcal{D}_{\Score}, \normed_1, \epsilon) \leq \Big( \frac{3 \, \corebound}{\epsilon} \Big)^{\prod \limits_{i=1}^{d} r_i}, \epsilon \in (0,1).
\end{align*}
\label{lemma:covering_sphere}
\end{lemma}
\noindent Similarly, to obtain a bound on $\covering(\mathcal{D}_{U_i}, \dist_{U_i}, \epsilon)$ for any $i\in [[d]]$, we employ the following lemma, which is proved in Appendix~\ref{app:CoveringNumber_matrix}.

\begin{lemma}
Define $\mathcal{D}_{U} := \{ U\in \mathbb{R}^{n\times r}: \|U(:,j)\|_2 \leq 1, \|U(:,j)\|_0 \leq s \text{ for all } j\in [[r]]\}$ with distance measure $\dist_{U}$, where $\dist_{U} (U^{(1)}, U^{(2)}) = \| U^{(1)} - U^{(2)} \|_{1,2}$ for any  $U^{(1)}, U^{(2)} \in \mathcal{D}_{U}$. Then the covering number of $\mathcal{D}_{U}$ with respect to the metric $\dist_{U}$ satisfies the bound
\begin{align*}
\covering(\mathcal{D}_{U}, \dist_{U}, \epsilon) \leq \Big( \frac{3n}{\epsilon} \Big)^{s r}, \epsilon \in (0,1).
\end{align*}
\label{lemma:covering_matrices}
\end{lemma}
\noindent Therefore, the bound in \eqref{bound:Cartesian} is evaluated using Lemma~\ref{lemma:covering_sphere} and Lemma~\ref{lemma:covering_matrices}.

Given a \revise{bound on} $\covering(\mathcal{D}_{P}, \dist_{P}, \epsilon)$ from \eqref{bound:Cartesian}, we are ready to derive a bound on the covering number of $\Class$ with respect to the metric $\dist_{\mathcal{G}}$. To this end, define a mapping $\Phi$ such that
\begin{align*}
\Phi(\Score, U_1, U_2,\ldots,U_d) = \Score \times_1 U_1 \times_2 U_2 \times_3 \cdots \times_d U_d
\end{align*}
where $(\Score, U_1, U_2,\ldots,U_d) \in \mathcal{D}_{P}$. Note from this definition that $\Phi:\mathcal{D}_{P} \rightarrow \Class$. We now employ the following lemma, which is formally proved in Sec.~\ref{app:lipschitz_constant}, to establish that this mapping $\Phi$ is Lipschitz with a Lipschitz constant of $\corebound \, (d+1)$.

\begin{lemma}
Consider $(\mathcal{D}_{P}, \dist_{P})$ to be the Cartesian product of metric spaces $(\mathcal{D}_{\Score}, \dist_{\Score})$, $(\mathcal{D}_{U_1}, \dist_{U_1})$, $(\mathcal{D}_{U_2}, \dist_{U_2})$, $\ldots$ , $(\mathcal{D}_{U_d}, \dist_{U_d})$, as defined in \eqref{eq:product_set} and \eqref{eq:product_metric}. Define mapping $\Phi$ such that
\begin{align*}
\Phi(\Score, U_1, U_2,\ldots,U_d) = \Score \times_1 U_1 \times_2 U_2 \times_3 \cdots \times_d U_d
\end{align*}
where $(\Score, U_1, U_2,\ldots,U_d) \in \mathcal{D}_{P}$. Further, define metric space $( \Class, \dist_{\mathcal{G}} )$, where $\Class$ is defined in \eqref{eq:model_main_2}, and $\dist_{\mathcal{G}} (\mathbf{G}^{(1)}, \mathbf{G}^{(2)}) = \| \mathbf{G}^{(1)} - \mathbf{G}^{(2)}\|_F$ for any $\mathbf{G}^{(1)}, \mathbf{G}^{(2)} \in \Class$. Then, given $P^{(1)}, P^{(2)}\in \mathcal{D}_{P}$, we have
\begin{align}
\dist_{\mathcal{G}} (  \Phi(P^{(1)}), \Phi(P^{(2)}) ) \leq \corebound (d+1) \dist_{P}(P^{(1)}, P^{(2)}) \text{.}
\label{eq:Lipschitzmapping_main}
\end{align}
\label{lemma:lipschitz_constant}
\end{lemma}
Finally, the application of Lemma~\ref{lemma:lipschitz_constant} with \eqref{bound:Cartesian} and Lemma~\ref{lemma:lipscitz_map}, where \eqref{bound:Cartesian} follows from Lemma~\ref{lemma:covering_cartesian}, establishes the statement of Lemma~\ref{lemma:low_rank_sparse_covering}.

\subsubsection{Deviation bound}\label{subsubsec:Deviation_bound}
Since $\deltarip = \sup \limits_{\Z \in \Class} \Big| \| \A( \Z) \|_2^2 - \mathbb{E} \big[ \| \A (\Z) \|_2^2 \big] \Big|$, we derive a probabilistic bound on the right hand side of this equality to evaluate the condition in \eqref{def:RIP}. To this end, we use techniques similar to those in~\cite{rauhut2017, krahmer2014suprema}. Specifically, define $\bm{\xi}$ to be a random vector in $\mathbb{R}^{n_1 \, n_2 \, \ldots n_d \, m}$ with independent entries from zero-mean, unit-variance, $subG(B)$ random variables. Further, let $\Z \in \Class$, and define $\Vz$ to be a matrix in $\mathbb{R}^{m \times n_1 \, n_2 \, \cdots \, n_d \, m}$ such that
\[ \Vz = \frac{1}{\sqrt{m}} \;\;  \mathbb{I}_{m} \otimes \Zvec^\top \]
where $\Zvec\in \mathbb{R}^{n_1 \, n_2 \, \cdots \, n_d \times 1}$ is the vectorized version of $\Z$. Then, we have the equivalence relationship $\A ( \Z ) \; = \;  \Vz \, \bm{\xi} $. For ease of notation, let us further define a set $\Mset := \{ \Vz: \Z \in \Class \}$. With this additional notation, we have $\deltarip = \sup \limits_{\M \in \Mset} \Big| \| \M \bm{\xi} \|_2^2 - \mathbb{E} \big[ \| \M \bm{\xi} \|_2^2 \big] \Big|$, and we apply the following theorem to obtain a deviation bound on the right hand side of this equality.

\begin{theorem} [\cite{krahmer2014suprema, rauhut2017}]
Let $\Mset_0$ be a set of matrices, and let $\bm{\xi_0}$ be a random vector with independent entries from zero-mean, unit-variance, $subG(\alpha_0)$ random variables. For the set $\Mset_0$, define
\begin{align*}
& d_F( \Mset_0 ) := \sup \limits_{\M \in \Mset_0} \|\M\|_F \text{,}
\;\;
d_{2\rightarrow 2}( \Mset_0 ) := \sup \limits_{\M \in \Mset_0} \|\M\|_{2} \text{,}
\nonumber\\
&\text{and } d_{4}( \Mset_0 ) := \sup \limits_{\M \in \Mset_0} \|\M\|_{S_4}
=  \sup \limits_{\M \in \Mset_0} \Big(  tr \big[ (\M^\top \M) ^2 \big] \Big)^{\frac{1}{4}} \text{.}
\end{align*}
Furthermore, let $\gamma_2(\Mset_0, \normed_2)$ be the Talagrand's $\gamma_2$-functional~\cite{talagrand2014upper}. Finally, set
\begin{align*}
& E_0 = \gamma_2(\Mset_0, \normed_2)
( \gamma_2(\Mset_0, \normed_2) + d_F(\Mset_0) ) + d_F(\Mset_0) d_{2 \rightarrow 2}(\Mset_0)
\\
& V_0 = d_4^2 (\Mset_0) \text{, and } U_0 = d_{2\rightarrow 2}^2 (\Mset_0) \text{.}
\end{align*}
Then, for $t>0$,
\begin{align*}
\mathbb{P}
\bigg( \sup \limits_{\M \in \Mset_0} \Big| \| \M \bm{\xi} \|_2^2 - \mathbb{E} \big[ \| \M \bm{\xi} \|_2^2 \big] \Big| \geq c_3 E_0 + t \bigg)
\leq 2 \exp \bigg(  - c_4 \min \bigg\{  \frac{t^2}{V_0^2}, \frac{t}{U_0} \bigg\}   \bigg) \text{,}
\end{align*}
where the positive constants $c_3$, $c_4$ depend on $\alpha_0$.
\label{th:deviationbound}
\end{theorem}

For the application of Theorem~\ref{th:deviationbound}, we need to evaluate bounds on the metrics $d_F( \Mset )$, $d_{2\rightarrow 2}( \Mset )$, $d_{4}( \Mset )$, and $\gamma_2(\Mset, \normed_2)$. However, the main analytical challenge in this application is evaluation of a bound on the Talagrand's $\gamma_2$-functional $\gamma_2(\Mset, \normed_2)$, which encompasses a geometric characterization of the metric space $(\Mset, \normed_2)$. We obtain a bound on the Talagrand's $\gamma_2$-functional using the following inequality~\cite{talagrand2014upper, rauhut2017}:
\begin{align}
\gamma_2(\Mset, \normed_{2}) \leq C \int_{0}^{d_{2\rightarrow 2}(\Mset)}
\; \sqrt{\log \covering (\Mset, \normed_2, \epsilon)}  \; d \epsilon \text{,}
\label{eq:TalagrandBoundMain}
\end{align}
where $C>0$ and $\covering (\Mset, \normed_2, u)$ denotes the covering number of the metric space ($\Mset$, $\normed_2$) with respect to the metric $\normed_2$. Thus, we employ the bound on covering number of $\Class$ from Lemma~\ref{lemma:low_rank_sparse_covering} to evaluate \eqref{eq:TalagrandBoundMain}, which enables us to obtain a bound on $\sup \limits_{\M \in \Mset} \Big| \| \M \bm{\xi} \|_2^2 - \mathbb{E} \big[ \| \M \bm{\xi} \|_2^2 \big] \Big|$ using Theorem~\ref{th:deviationbound}. A formal proof of Theorem~\ref{th:subGaussian} follows in Sec.~\ref{app:proof_subGaussian}.

\section{Numerical Experiments}\label{sec:Experiments}
In this section, we perform experiments on synthetic and real-world data to analyze the performance of the proposed TPGD method (Algorithm~\ref{algo:TPG}), which, as explained before, is a tensor variant of the projected gradient descent (PGD) method. We compare TPGD with learning methods based on ($i$) vectorization of the parameter tensor, ($ii$) imposition of low Tucker-rank, and ($iii$) imposition of low CP-rank~\cite{kolda2009tensor} on the parameter tensor $\B$. To analyze linear vectorization-based methods, we employ LASSO~\cite{tibshirani1996regression} and linear support vector machine regression (SVR)~\cite{ho2012large}. To analyze imposition of low Tucker-rank and low CP-rank, we employ Tucker-rank and CP-rank variants of the tensor projected gradient descent method, respectively. Specifically, in the first variant, projection is performed onto a set of low Tucker-rank tensors~\cite{rauhut2017}, and we call this method PGD-Tucker. In the second variant, projection is performed onto a set of low CP-rank tensors~\cite{zhou2013tensor}, and we call this method PGD-CP. Thus, we draw comparisons of TPGD (Algorithm~\ref{algo:TPG}) with LASSO, SVR, PGD-Tucker, and PGD-CP.

Some relevant implementation details for these learning methods are as follows. For computation of the projection step $\Hsr$ in Algorithm \ref{algo:TPG}, we employ Algorithm~\ref{algo:TPGprojection}, within which we employ the inverse power method from~\cite{hein2010inverse} for computation of Step~3. For computation of the projection steps in the Tucker rank (PGD-Tucker) and the CP rank (PGD-CP) based methods, we employ the tensor toolboxes in~\cite{vervliettensorlab} and \cite{TTB_Software}, respectively. Finally, we employ MATLAB's built-in fitrlinear function~\cite{MATLAB2018} for implementing LASSO and SVR methods.

\subsection{Synthetic Experiments}\label{sec:Synthetic_Exp}
For synthetic-data experiments, we generate the $\rank$-rank and $\sparsity$-sparse tensor $\W\in \mathbb{R} ^ {n_1 \times n_2 \times \dots \times n_d }$ in \eqref{eq:Measurement} as follows. We set $d=3$, $n_1=50$, $n_2=50$, $n_3=30$, and in \eqref{eq:Model}, we set $s_1=6$, $s_2=6$, $s_3=4$, with $r_1=3$, $r_2=3$, and $r_3=3$. For each $j\in[[d]]$, we generate the column vector $\U_j(:,i)$, for each $i\in[[r]]$, such that $\|\U_j(:,i)\|_0\leq s_j$. The locations of the $s_j$ non-zero entries in $\U_j(:,i)$ are chosen uniformly at random from $[[n_j]]$. Setting $a=0.5$, we sample the non-zero entries in $\U_j(:,i)$ from $(-1)^u (a+|z|)$, where $u$ is drawn from a Bernoulli distribution with parameter $0.5$ and $z$ is drawn from a standard Gaussian distribution, i.e., $\text{Gaussian}(0,1)$. Finally, to generate the parameter tensor $\W$, the entries of the core tensor $\Score$ are sampled from a uniform distribution with parameters $0$ and $1$, and the tensor $\W$ is generated as in \eqref{eq:Model}. To generate the response vector $\y$, the tensors $\{ \X_i \}_{i=1}^{m}$ are generated such that their entries are i.i.d. $\text{Gaussian}(0,1/m)$, the noise vector $\noise$ is sampled from $\text{Gaussian}(0,\sigma_{z}^2 I)$, and then, the response vector $\y$ is generated as in \eqref{eq:Measurement}.

\begin{figure}
\centering
\subfloat[]
{ \includegraphics[draft=false,scale=0.24]{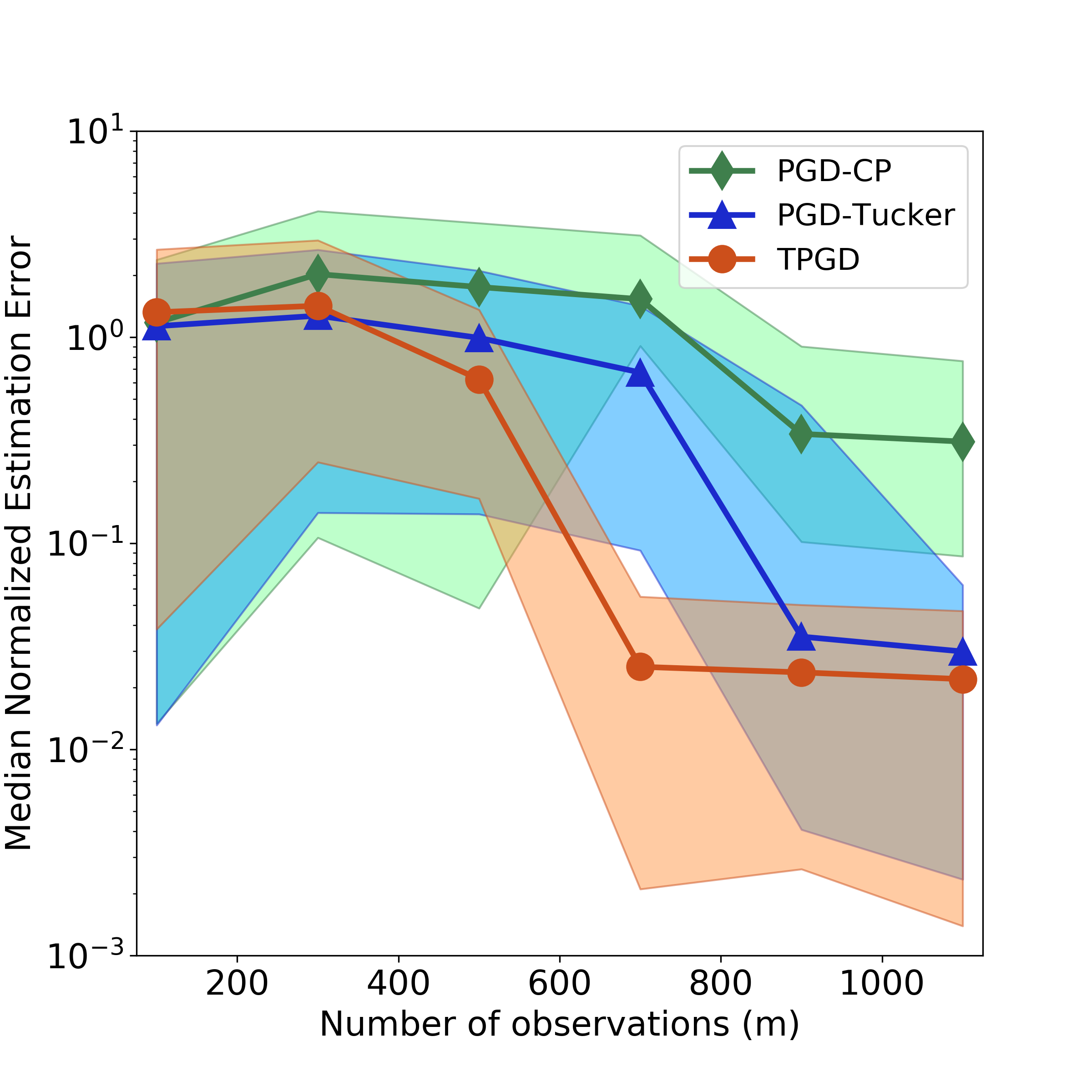}
\label{fig:synthetic_median_01}}
\subfloat[]
{ \includegraphics[draft=false,scale=0.24]{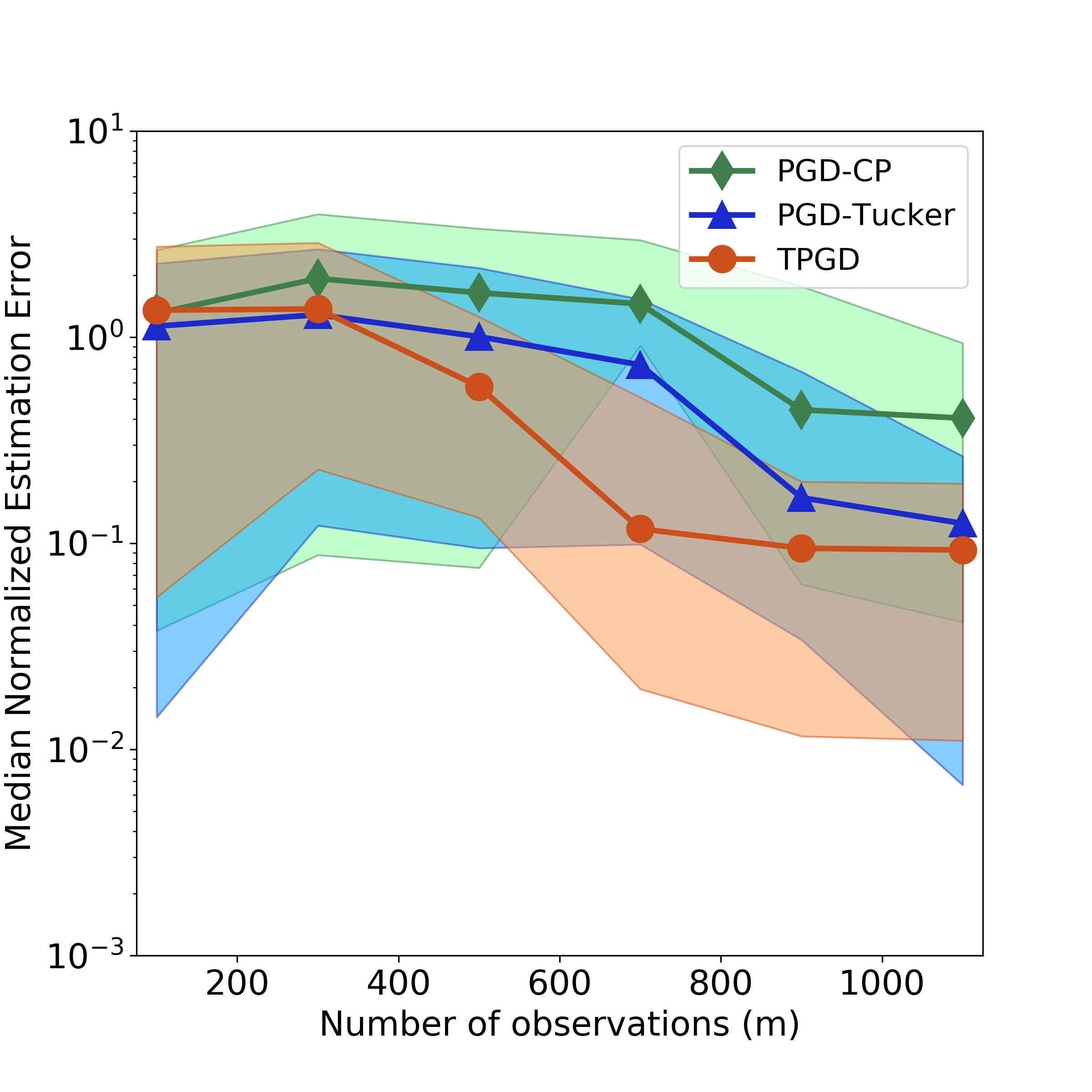}
\label{fig:synthetic_median_04}}
\subfloat[]
{ \includegraphics[draft=false,scale=0.24]{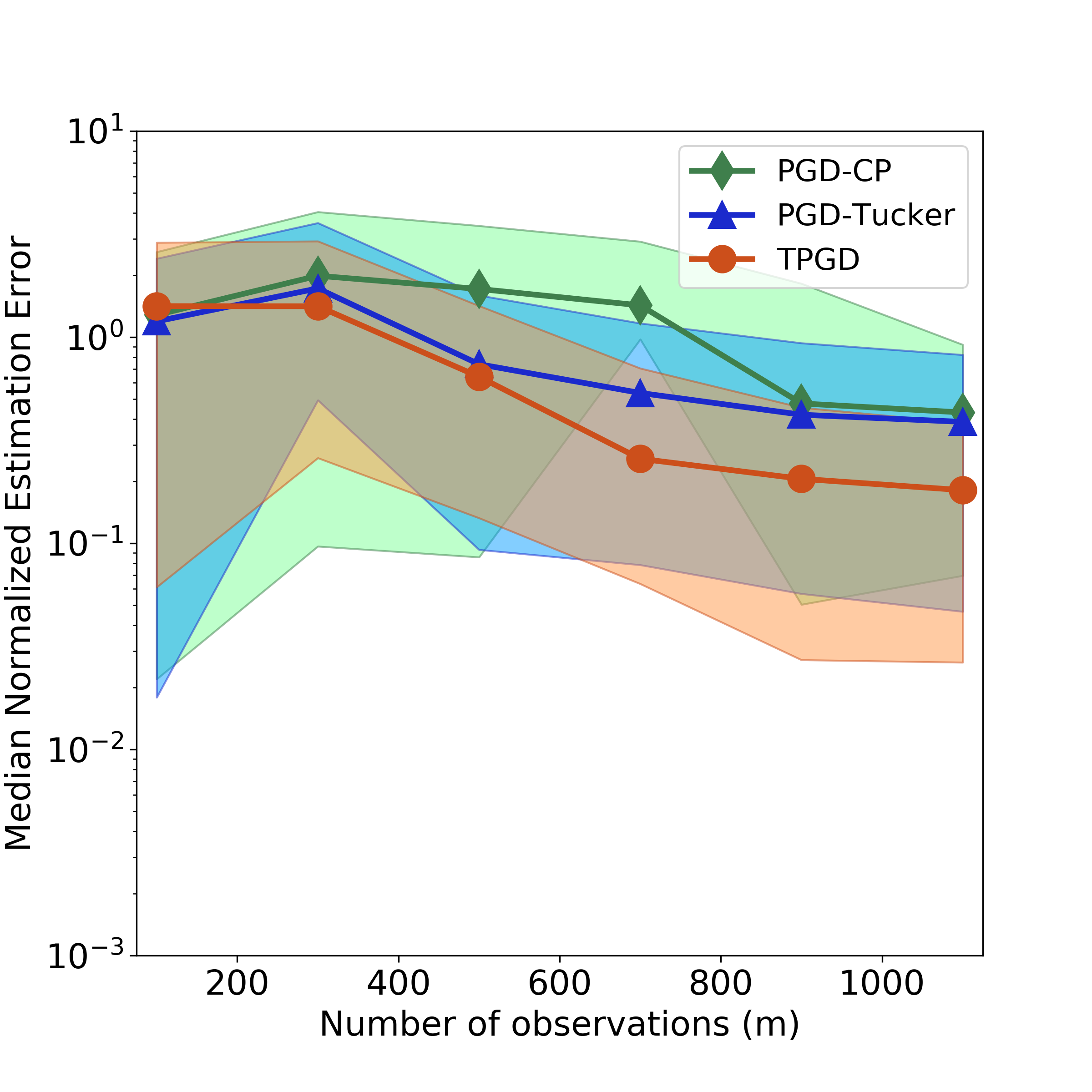}
\label{fig:synthetic_median_07}}

\subfloat[]
{ \includegraphics[draft=false,scale=0.3]{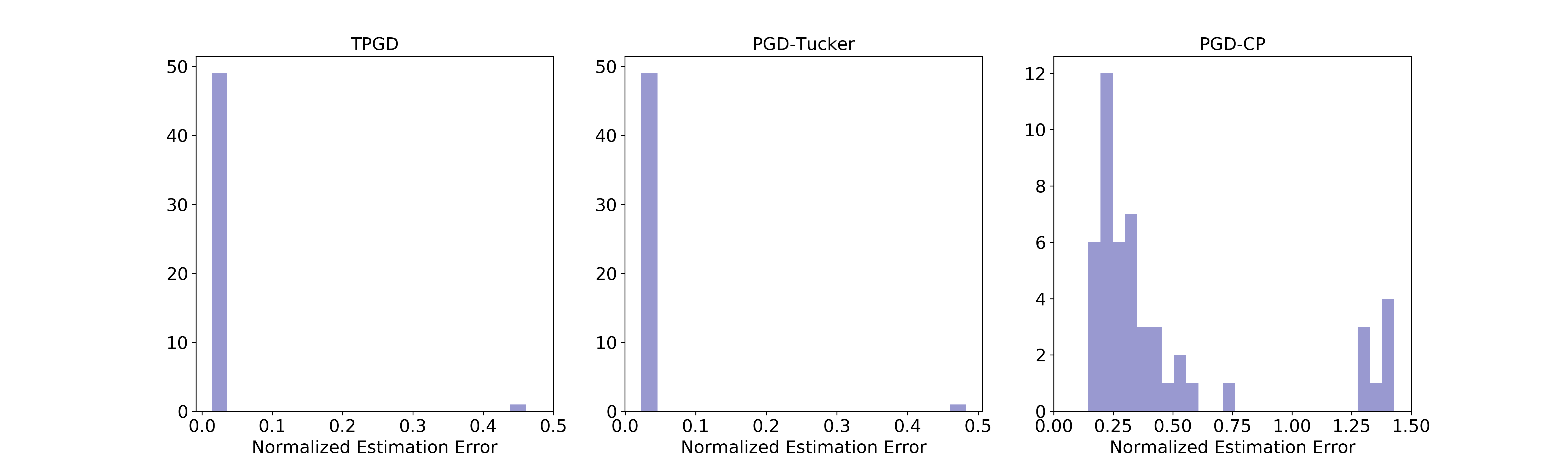}
\label{fig:histogram_failure}}

\centering
\subfloat[]
{ \includegraphics[draft=false,scale=0.3]{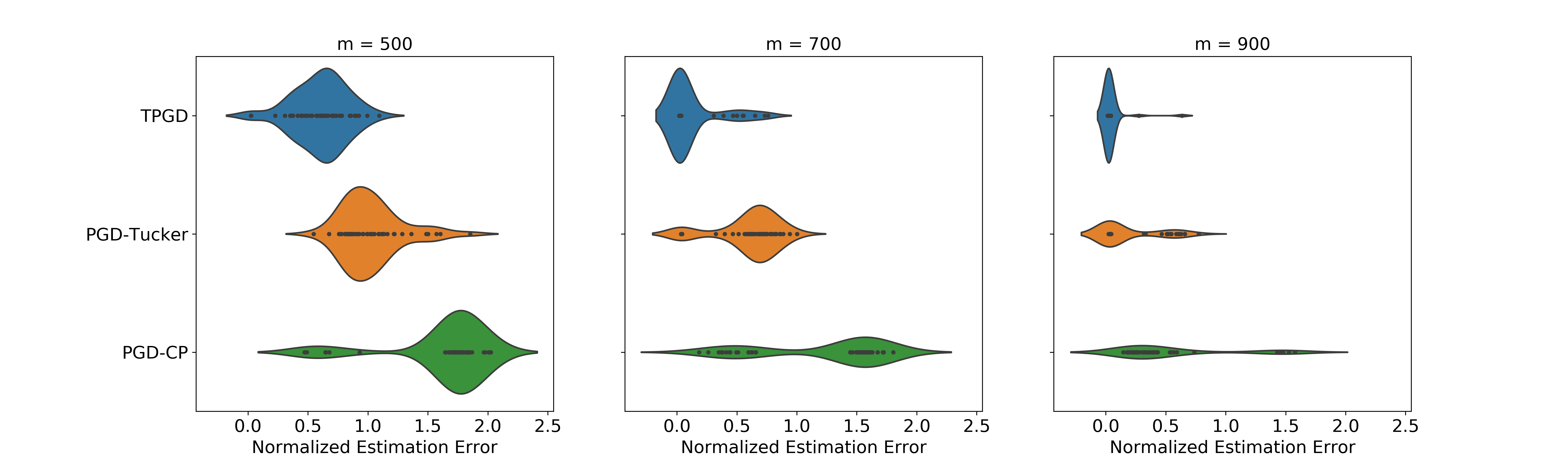}
\label{fig:violin_plot}}
\caption{Comparison of TPGD with PGD-Tucker and PGD-CP over synthetic data for (a) $\sigma_z=0.1$, (b) $\sigma_z=0.4$, and (c) $\sigma_z=0.7$. For each value of $m$, the markers in (a)--(c) correspond to the median estimation error over $50$ experiments; whereas, the shaded region for each marker pertains to the 25th and the 75th percentiles of the estimation error. Note that we report median error since an occasional failure in recovering the parameter tensor may lead to a spike in mean error. \revise{In order to analyze the failure rates, (d) shows histograms of estimation errors for $\sigma_z = 0.1$ and $m=1100$ for the three methods. In addition, (e) shows empirical distributions of the estimation errors for $\sigma_z = 0.1$ in terms of violin plots, corresponding to the values of $m$ around the phase transition region of TPGD.}}
\label{fig:synthetic_median}
\end{figure}

The aforementioned experiment is performed for various values of $m$, repeating each experiment for increasing value of $\sigma_z$ to analyze the impact of increasing noise power.  For each value of $m$ and $\sigma_z$, ($i$) the parameter tensor $\W$, the linear map $\A$, and the response vector $\y$ are generated as explained above, and ($ii$) a parameter estimate $\Walgo$ is computed using each of the learning methods. The algorithmic parameters for each of the learning methods are set using separate validation experiments. The performance of each learning method is characterized using the normalized estimation error, which is defined as $\frac{\| \W - \Walgo \|_F}{\| \W \|_F}$. For each value of $m$ and $\sigma_z$, this experimental procedure is repeated $50$ times, and the median estimation error is reported in Fig.~\ref{fig:synthetic_median_01}--Fig.~\ref{fig:synthetic_median_07}, along with the 25th and the 75th percentile of estimation error. \revise{Further, in order to compare the `failure' rates of different methods, which correspond to relatively large estimation errors, we plot the histograms of estimation errors for $\sigma_z = 0.1$ and $m=1100$ in Fig.~\ref{fig:histogram_failure} for the three methods of TPGD, PGD-Tucker, and PGD-CP. It can be seen from these histograms that the failure rates of both TPGD and PGD-Tucker are quite small. Finally, in order to characterize the empirical distributions of the estimation errors over all experiments, we also report violin plots of the estimation errors for the three methods in Fig.~\ref{fig:violin_plot} for values of $m$ around the phase transition region for the TPGD algorithm.} Note that LASSO and SVR perform considerably \revise{worse} than the other learning methods; thus, they are not included in Fig.~\ref{fig:synthetic_median} for clarity of plots.

We gain two interesting insights from Fig.~\ref{fig:synthetic_median}. First, the plots show that the projection step $\Hsr$ in Algorithm~\ref{algo:TPG} (TPGD method) can be computed accurately enough by Algorithm~\ref{algo:TPGprojection}, enabling the TPGD method to achieve better sample complexity compared with the other learning methods, by exploiting the low-rank and sparse structure in the parameter tensor. In other words, despite the lack of theoretical guarantees for Algorithm~\ref{algo:TPGprojection}, it can be employed in practice to compute the projection step \eqref{eq:Hsr} in Step~5, Algorithm~\ref{algo:TPG}. Second, comparing Fig.~\ref{fig:synthetic_median_01}, Fig.~\ref{fig:synthetic_median_04}, and Fig.~\ref{fig:synthetic_median_07}, we see that as the noise power decreases, the accuracy of the solution of TPGD increases. This is also reflected in the statement of Theorem \ref{th:main_result}: the lower the noise power, the more accurate the solution of TPGD, and vice versa.

\revise{Finally, we numerically investigate the computational complexity of our specific implementations of TPGD, PGD-Tucker, and PGD-CP for various values of $n_1$, $n_2$, and $n_3$. To this end, we repeat the aforementioned experiment for $m=500$, $\sigma_z = 0.1$, and $n_1=n_2=n_3=:n$, except that we fix the maximum number of iterations to $100$ and we vary $n$ to have values of ($i$) $n=10$, ($ii$) $n=20$, and ($iii$) $n=40$. For each of these values of $n$, we report the per-iteration computational time of each method in Table~\ref{tab:time}. It can be seen from this table that the mean computational time is comparable for the three tensor-based methods.}

\begin{table}[h]
\caption{\revise{Per-iteration computational time (in seconds) of TPGD, PGD-Tucker, and PGD-CP implementations for ($i$) $n=10$, ($ii$) $n=20$, and ($iii$) $n=40$, reported as an average over the $50$ experiments. The variance is also reported in parenthesis in each cell of the table.}}
\label{tab:time}
\centering
\revise{
\begin{tabular}{ |p{3.1cm}||p{2.2cm}|p{2.2cm}|p{2.2cm}|  }
 \hline
   & $n=10$  & $n=20$ & $n=40$ \\
 \hline
TPGD          & 0.0096 (1e-6)        & 0.031 (5e-7) & 0.32 (8e-5)  \\
PGD-Tucker         & 0.0021 (1e-7)  & 0.021 (5e-7) & 0.24 (4e-5) \\
PGD-CP          & 0.047 (3e-4)    & 0.060 (2e-4) & 0.27 (8e-5)
\\
\hline
\end{tabular}
}
\end{table}

\subsection{Neuroimaging Data Analysis}\label{sec:Real_Exp}
We also analyze the performance of TPGD for predicting attention deficit hyperactivity disorder (ADHD) diagnosis, using a preprocessed repository of ADHD-200 fMRI images~\cite{milham2012adhd} from the Donders Institute (NeuroImage), the Kennedy Krieger Institute (KKI), and the NYU Child Study Center (NYU). Specifically, we use preprocessed brain maps of fractional amplitude of low-frequency fluctuations (fALFF)~\cite{zou2008improved} that were obtained using the Athena pipeline~\cite{bellec2017neuro}. Note that fALFF is defined as the ratio of power within the low-frequency range (0.01--0.1 Hz) to that of the entire frequency range and as such it characterizes the intensity of spontaneous brain activity. Altered levels of fALFF have been reported in a sample of children with ADHD relative to controls \cite{yang2011abnormal}, so fALFF brain maps form a useful feature space for predicting ADHD diagnosis.

The train data consists of fALFF brain maps for individuals pertaining to NeuroImage, KKI, and NYU. For each of these imaging sites, each individual's fALFF map forms a third-order tensor $\X_i \in \mathbb{R}^{49 \times 58 \times 47}$, and the ADHD diagnosis $y_i$ ($1=$ ADHD, $0=$ normal control) forms the response, where $i\in [[m]]$ and $m$ is the number of train samples. In our experiments, we have $m=39$ for NeuroImage \revise{(ADHD = $17$, control = $22$)}, $m=78$ for KKI \revise{(ADHD = $20$, control = $58$)}, and $m=188$ for NYU \revise{(ADHD = $97$, control = $91$)}, and we learn a regression model for each site, independently. Given fALFF maps $\{ \X_i \}_{i=1}^{m}$ and responses $\{ y_i \}_{i=1}^{m}$ for each site, the task of learning the regression model in \eqref{eq:System1} is equivalent to learning the parameter tensor $\W$. We estimate the unknown parameter tensor using TPGD, PGD-Tucker, PGD-CP, LASSO, and SVR.

\begin{table}[h]
\caption{Comparison of TPGD with PGD-Tucker, PGD-CP, LASSO, and SVR for predicting diagnosis of test subjects corresponding to (a) Donders Institute (NeuroImage), (b) Kennedy Krieger Institute (KKI), and (c) New York University Child Study Center (NYU), respectively.}
\label{table:ADHD2000}
\centering
\textbf{ (a) The Donders Institute (NeuroImage)}
\begin{tabular}{ |p{3.1cm}||p{1.8cm}|p{2.2cm}| p{1.8cm}| p{1.8cm}|p{1.8cm}| }
 \hline
   & TPGD  & PGD-Tucker & PGD-CP & LASSO & SVR \\
 \hline
Specificity           & 0.68 & 0.57 & 0.57 & 1 &  0.89 \\
Sensitivity          & 0.73 & 0.45 & 0.64 & 0.18 &  0.36 \\
\textbf{Harmonic mean}   & \textbf{0.70} & 0.50 & 0.60 & 0.31 &  0.51 \\
\hline
\end{tabular}
\textbf{ (b) Kennedy Krieger Institute (KKI) }
\begin{tabular}{ |p{3.1cm}||p{1.8cm}|p{2.2cm}| p{1.8cm}| p{1.8cm}|p{1.8cm}| }
 \hline
   & TPGD  & PGD-Tucker & PGD-CP & LASSO & SVR\\
 \hline
Specificity           & 0.63 & 0.50 & 0.50 & 1 &  1 \\
Sensitivity         & 0.67 & 0.33 & 0.33 & 0 &  0 \\
\textbf{Harmonic mean}   & \textbf{0.65} & 0.40 & 0.40 & 0 &  0 \\
\hline
\end{tabular}
\textbf{(c) New York University Child Study Center (NYU)}
\begin{tabular}{ |p{3.1cm}||p{1.8cm}|p{2.2cm}| p{1.8cm}| p{1.8cm}|p{1.8cm}| }
 \hline
   & TPGD  & PGD-Tucker & PGD-CP & LASSO & SVR \\
 \hline
Specificity           & 0.58 & 0.67 & 0.67 & 0.42  & 0.17 \\
Sensitivity           & 0.52 & 0.52 & 0.48 & 0.55 & 0.59 \\
\textbf{Harmonic mean}   & 0.55 & \textbf{0.59} & 0.56 & 0.48 & 0.26 \\
\hline
\end{tabular}
\end{table}

To analyze the performance of these learning methods, we employ separately provided test datasets for NeuroImage, KKI, and NYU, pertaining to fALFF maps of $25$, $11$, and $41$ test subjects, respectively. To analyze the performance for each method, we use the estimate of $\W$ to compute the responses for the test subjects using \eqref{eq:System1}. If the computed response is more than $0.5$ for a test subject, the subject is labeled with ADHD and vice versa. To evaluate the predictive power of each method using test data, we use the notion of ($i$) specificity, which is the ratio of subjects not diagnosed with ADHD that are correctly labeled as normal controls, and ($ii$) sensitivity, which is the ratio of subjects diagnosed with ADHD that are correctly labeled with ADHD. The explained experimental procedure is repeated $50$ times for each method and imaging site, and the median results on test data are reported in Table~\ref{table:ADHD2000}, along with the harmonic mean of reported specificity and sensitivity. The TPGD method tends to perform well in the low sample size regime, given that it provides the highest harmonic mean on test data for the NeuroImage and the KKI sites, respectively. \revise{Moreover, we observe that vectorization-based methods of LASSO and SVR perform poorly on the KKI test dataset, which entails a challenging prediction task because of the high class imbalance in the KKI train dataset.} For the NYU imaging site, the PGD-Tucker method tends to work best; however, the performance of PGD-CP and TPGD methods is not much \revise{worse} either. \revise{The slightly worse performance of TPGD compared to PGD-Tucker is attributable to differences in implementations of the projection steps for each method.}

\section{Conclusion}\label{sec:Conclusion}
In this work, we studied a tensor-structured linear regression model, with simultaneous imposition of a sparse and low Tucker-rank structure on the parameter tensor. We formulated the parameter estimation problem as a non-convex program, and then we proposed a projected gradient descent-based method to solve it. In our analysis, we provided mathematical guarantees for the proposed method based on the restricted isometry property. Furthermore, we evaluated the property for the case of sub-Gaussian predictors, characterizing the sample complexity of parameter estimation in the process. Finally, in our experiments with real-world data, we demonstrated that the simultaneously-structured tensor regression model is not restrictive, and it can be effectively employed for neuroimaging data analysis.

\appendix

\section{Proof of Lemma~\ref{lemma:add_tensors}}\label{app:proof_add_tensors}
Since $\Z_a \in \Class$, it can be expressed as
\begin{align*}
\Z_a = \Score_a \times_1 \U_{a,1} \times_2 \U_{a,2} \cdots \times_d \U_{a,d} \text{,}
\end{align*}
where $\Score_a \in \mathbb{R} ^ {r_1 \times r_2 \times \dots \times r_d }$ such that $\|\Score_a\|_1 \leq \corebound$, and $\U_{a,i} \in \mathbb{R} ^ {n_i \times r_i}$, with $\| \U_{a,i}(:,j) \|_0\leq s_i$, $\forall i\in [[d]]$, $j\in [[r_i]]$. Similarly, since $\Z_b \in \Class$, it can be expressed as
\begin{align*}
\Z_b = \Score_b \times_1 \U_{b,1} \times_2 \U_{b,2} \cdots \times_d \U_{b,d} \text{,}
\end{align*}
where $\Score_b \in \mathbb{R} ^ {r_1 \times r_2 \times \dots \times r_d }$ such that $\|\Score_b\|_1 \leq \corebound$, and $\U_{b,i} \in \mathbb{R} ^ {n_i \times r_i}$, with $\| \U_{b,i}(:,j) \|_0\leq s_i$, $\forall i\in [[d]]$, $j\in [[r_i]]$. Let $\Z_c = \gamma_a \Z_a + \gamma_b \Z_c$, where $\gamma_a\in \mathbb{R}$, $\gamma_b\in \mathbb{R}$, so that $\Z_c$ is some linear combination of $\Z_a$ and $\Z_b$. Define the Cartesian product $\mathcal{D}_P:=[[r_1]] \times [[r_2]] \times \cdots \times [[r_d]]$. Using the definition of $\mathcal{D}_P$, define $\Score_c \in \mathbb{R}^{2 r_1 \times 2 r_2\times \ldots \times 2 r_d}$ where

\[
\Score_c (i_1, i_2, \ldots, i_d) =
  \begin{cases}
    \gamma_a \, \Score_a (i_1, i_2, \ldots, i_d)  & : (i_1, i_2, \ldots, i_d) \in \mathcal{D}_P  \\
    \gamma_b \, \Score_b (i_1, i_2, \ldots, i_d)  &  : (i_1 - r_1, i_2 - r_2, \ldots, i_d - r_d) \in \mathcal{D}_P \\
        0  &  :\text{otherwise}
  \end{cases}
\]
for $(i_1, i_2, \ldots, i_d) \in [[2 r_1]] \times [[2 r_2]] \times \cdots \times [[2 r_d]]$. Note that $\|\Score_c\|_1 = \| \gamma_a \Score_a\|_1 + \| \gamma_b \Score_b\|_1 \leq  (|\gamma_a| + |\gamma_b|) \corebound$. Furthermore, for $i \in [[d]]$, define $\U_{c,i} \in \mathbb{R}^{n_i \times 2 r_i}$ such that $\U_{c,i} := [ \U_{a,i} \;\;  \U_{b,i}  ] $. Finally, with these definitions, $\Z_c$ can be expressed as
\begin{align*}
\Z_c = \Score_c \times_1 \U_{c,1} \times_2 \U_{c,2} \cdots \times_d \U_{c,d}\text{,}
\end{align*}
where $\Score_c \in \mathbb{R}^{2 r_1 \times 2 r_2\times \ldots \times 2 r_d}$ such that $\|\Score_c\|_1 \leq  (|\gamma_a| + |\gamma_b|) \corebound$, and $\U_{c,i} \in \mathbb{R}^{n_i \times 2 r_i}$ such that $\| \U_{c,i}(:,j) \|_0\leq s_i$, for all $i\in [[d]]$, $j\in [[2 r_i]]$. Therefore, $\Z_c$ is a member of the set $\mathcal{G}_{2\rank,\sparsity, \kappa }$, where $\kappa = (|\gamma_a| + |\gamma_b|) \corebound$.
\qed

\section{Proof of Theorem~\ref{th:main_result}}
\label{app:main_result}
Let $\Loss(\Z) := \| \y - \A(\Z) \|_2^2$ be the loss function for any $\Z\in \mathbb{R}^{n_1\times n_2 \times \ldots \times n_d}$. Then, we have
\begin{align}
\Loss(\Wkplus) - \Loss(\Wk)
&= \| \y - \A(\Wkplus) \|_2^2 - \| \y - \A(\Wk) \|_2^2
\nonumber\\
& = \| \A(\Wkplus) \|_2^2 - \|\A(\Wk) \|_2^2 - 2 \langle \y, \A(\Wkplus - \Wk) \rangle
\nonumber\\
& = \| \A(\Wkplus) \|_2^2 + \|\A(\Wk) \|_2^2 - 2 \|\A(\Wk) \|_2^2 - 2 \langle \y, \A(\Wkplus - \Wk) \rangle
\nonumber\\
& = \| \A(\Wkplus - \Wk) \|_2^2 + 2 \langle \A(\Wk) , \A(\Wkplus) \rangle - 2 \langle \A(\Wk) , \A(\Wk) \rangle
\nonumber\\
& - 2 \langle \y, \A(\Wkplus - \Wk) \rangle
\nonumber\\
&= \| \A(\Wkplus - \Wk) \|_2^2 + 2 \langle \A(\Wk) - \y, \A(\Wkplus - \Wk)  \rangle
\nonumber\\
&= \| \A(\Wkplus - \Wk) \|_2^2 + 2 \langle \Aconjugate (\A(\Wk) - \y), \Wkplus - \Wk  \rangle
\nonumber\\
& \leq (1 + \deltariptwo) \| \Wkplus - \Wk \|_F^2 + 2 \langle \Aconjugate (\A(\Wk) - \y), \Wkplus - \Wk  \rangle \text{,}
\label{eq:main1}
\end{align}
where the last inequality follows from application of Definition \ref{def:RIP} with Lemma~\ref{lemma:add_tensors}.

For any $\Z\in \mathbb{R}^{n_1\times n_2 \times \ldots \times n_d}$, define
\begin{align}
\g(\Z)
&:= (1+\deltariptwo) \| \Z - \Wk \|_F^2 + 2 \langle \Aconjugate (\A(\Wk) - \y), \Z - \Wk  \rangle
\nonumber\\
& \eqa (1+\deltariptwo) \| \Z - \Wtildek + \mu \Aconjugate( \y - \A (\Wk)) \|_F^2
\nonumber\\
& + 2 \langle \Aconjugate (\A(\Wk) - \y), \Z - \Wtildek + \mu \Aconjugate( \y - \A (\Wk))  \rangle
\nonumber\\
& \eqb (1+\deltariptwo) \| \Z - \Wtildek \|_F^2 - \frac{1}{1 + \deltariptwo} \| \Aconjugate (\y - \A (\Wk)) \|_F^2 \text{,}
\label{eq:g(Z)}
\end{align}
where (a) follows by substituting $\Wk = \Wtildek + \mu \Aconjugate( \A (\Wk) - \y )$ and (b) follows by substituting $\mu = \frac{1}{1 + \deltariptwo}$. Then, since $\|\Wkplus - \Wtildek\|_F \leq \| \W - \Wtildek\|_F$, which follows from $\Wkplus = \Hsr(\Wtildek)$, we have $\g(\Wkplus)\leq \g(\W)$. Using $\g(\Wkplus)\leq \g(\W)$ with \eqref{eq:main1}, we obtain
\begin{align}
\Loss(\Wkplus) - \Loss(\Wk)
&\leq (1+\deltariptwo) \| \W - \Wk\|_F^2 + 2 \langle \Aconjugate (\A(\Wk) - \y), \W - \Wk  \rangle
\nonumber\\
&= 2 \deltariptwo \| \W - \Wk\|_F^2 + (1-\deltariptwo) \| \W - \Wk\|_F^2
\nonumber\\
&\quad+ 2 \langle \Aconjugate (\A(\Wk) - \y), \W - \Wk  \rangle
\nonumber\\
& \leq 2 \deltariptwo \| \W - \Wk\|_F^2 + \| \A(\W - \Wk) \|_2^2
\nonumber\\
&\quad+ 2 \langle \Aconjugate (\A(\Wk) - \y), \W - \Wk  \rangle
\nonumber\\
& = 2 \deltariptwo \| \W - \Wk\|_F^2 + \| \A(\W - \Wk) \|_2^2
\nonumber\\
& \quad+ 2 \langle \A(\Wk), \A(\W - \Wk)  \rangle - 2 \langle \y, \A(\W - \Wk)  \rangle
\nonumber\\
& = 2 \deltariptwo \| \W - \Wk\|_F^2 + \| \A (\W) \|_2^2 - \| \A (\Wk) \|_2^2  - 2 \langle \y, \A(\W - \Wk)  \rangle
\nonumber\\
& = 2 \deltariptwo \| \W - \Wk\|_F^2 + \| \y - \A (\W) \|_2^2 - \| \y - \A (\Wk) \|_2^2
\nonumber\\
& \leq \frac{2 \deltariptwo}{1-\deltariptwo} \| \A(\W - \Wk)\|_2^2 + \Loss(\W) - \Loss(\Wk) \text{,}
\end{align}
where the last two inequalities follow from application of Definition \ref{def:RIP} with Lemma~\ref{lemma:add_tensors}. Thus, we have
\begin{align}
\Loss(\Wkplus) \leq \frac{2 \deltariptwo}{1-\deltariptwo} \| \A(\W - \Wk)\|_2^2  + \Loss(\W).
\label{eq:main2}
\end{align}
Using $\A(\W) = \y - \noise$, we have
\begin{align}
\|\A(\W-\Wk)\|_2^2
= \| \y - \A(\Wk) - \noise \|_2^2 \leq  2 (\| \y - \A(\Wk) \|_2^2 + \|\noise \|_2^2) =  2 \big( \Loss(\Wk) + \|\noise \|_2^2\big) \text{,}
\label{eq:main3}
\end{align}
where the inequality follows since $(u+v)^2 \leq 2 (u^2 + v^2)$ for all $u,v\in \mathbb{R}$.
Using \eqref{eq:main2} with \eqref{eq:main3}, and observing $\Loss(\W)=\|\noise\|_2^2$, we obtain
\begin{align}
\Loss(\Wkplus)
&\leq \frac{4 \deltariptwo}{1 - \deltariptwo} \Big(\Loss(\Wk) + \|\noise\|_2^2 \Big) + \|\noise\|_2^2
\nonumber\\
&= \frac{4 \deltariptwo}{1 - \deltariptwo} \Loss(\Wk) + \Big(1+\frac{4 \deltariptwo}{1 - \deltariptwo}\Big) \|\noise\|_2^2 \text{.}
\end{align}
Using $\deltariptwo<\frac{\gamma}{4+\gamma}$, $\gamma<1$, and $b=\frac{1+3\deltariptwo}{1-\deltariptwo}$ yields
\begin{align}
\Loss(\Wkplus)\leq \gamma \Loss(\Wk) + b \|\noise\|_2^2 \text{.}
\end{align}
Iterative application of this inequality leads to
\begin{align}\label{eqn:lbk}
\Loss(\Wk)\leq \gamma^{k} \Loss(\W^{0}) + \frac{b}{1-\gamma} \|\noise\|_2^2
\end{align}
for all $k\geq 1$.

\revise{
Next, using Definition \ref{def:RIP} with Lemma~\ref{lemma:add_tensors}, we obtain
\begin{align}
\|\W^{k} - \W\|_F^2
&\leq \frac{1}{1-\deltariptwo} \| \A(\W^k - \W) \|_2^2 \leq \frac{2}{1-\deltariptwo} \Big( \Loss(\W^k) + \|\noise\|_2^2 \Big),
\end{align}
where the last inequality follows from \eqref{eq:main3}. Finally, using \eqref{eqn:lbk}, we get
\begin{align}
\|\W^{k} - \W\|_F^2
&\leq \frac{2}{1 - \deltariptwo}\Big(\gamma^{k} \Loss(\W^{0}) + \frac{b}{1-\gamma} \|\noise\|_2^2 + \|\noise\|_2^2\Big)\nonumber\\
&= \frac{2 \gamma^k}{1-\deltariptwo}\mathcal{L}(\mathbf{B}^0)+\frac{2\|\noise\|_2^2}{1-\deltariptwo}\Big(1+\frac{b}{1-\gamma}\Big).
\end{align}
}
\qed

\section{Proof of Lemma~\ref{lemma:low_rank_sparse_covering}}\label{app:low_rank_sparse_covering}
Recall the metric space $(\mathcal{D}_{\Score}, \dist_{\Score})$, where $\mathcal{D}_{\Score} := \{ \Score \in \mathbb{R}^{r_1 \times r_2 \times \cdots \times r_d}: \|\Score\|_1 \leq \tau \}$ and $\dist_{\Score}(\Score^{(1)},\Score^{(2)}) := \frac{1}{\corebound} \| \Score^{(1)} - \Score^{(2)} \|_1$ for any $\Score^{(1)}, \Score^{(2)} \in \mathcal{D}_{\Score}$. Using Lemma~\ref{lemma:covering_sphere}, the covering number of $\mathcal{D}_{\Score}$ with respect to the metric $\dist_{\Score}$ satisfies the bound
\begin{align*}
\covering(\mathcal{D}_{\Score}, \dist_{\Score}, \epsilon) \leq \Big( \frac{3}{\epsilon} \Big)^{\prod \limits_{i=1}^{d} r_i}, \epsilon \in (0,1).
\end{align*}

\noindent Further, recall the metric space $(\mathcal{D}_{U_i}, \dist_{U_i})$, where $\mathcal{D}_{U_i} := \{ U\in \mathbb{R}^{n_i \times r_i}: \|U(:,j)\|_2 \leq 1, \|U(:,j)\|_0 \leq s_i, j\in [[r_i]]\}$ and $\dist_{U_i} (U_i^{(1)}, U_i^{(2)}) := \| U_i^{(1)} - U_i^{(2)} \|_{1,2}$ for any  $U_i^{(1)}, U_i^{(2)} \in \mathcal{D}_{U_i}$, $i\in [[d]]$. Using Lemma~\ref{lemma:covering_matrices}, the covering number of $\mathcal{D}_{U_i}$ with respect to the metric $\dist_{U_i}$ satisfies the bound
\begin{align*}
\covering(\mathcal{D}_{U_i}, \dist_{U_i}, \epsilon) \leq \Big( \frac{3 \bar{n} }{\epsilon} \Big)^{s_i r_i}, \epsilon \in (0,1),
\end{align*}
for any $i\in [[d]]$. Next, recall the metric space $(\mathcal{D}_{P}, \dist_{P})$, where
\begin{align*}
\mathcal{D}_{P} := \mathcal{D}_{\Score} \times \mathcal{D}_{U_1} \times \mathcal{D}_{U_2} \times \cdots \times \mathcal{D}_{U_d} \text{, and}
\end{align*}
\begin{align*}
\dist_{P}(P^{(1)}, P^{(2)}) = \max \big\{  \max \limits_{i\in [[d]]} \{  \dist_{U_i} (U_i^{(1)}, U_i^{(2)}) \}, \dist_{\Score}(\Score^{(1)},\Score^{(2)}) \big\} \text{,}
\end{align*}
such that $P^{(1)}, P^{(2)}\in \mathcal{D}_{P}$, $\Score^{(1)}, \Score^{(2)} \in \mathcal{D}_{\Score}$, and $U_i^{(1)}, U_i^{(2)} \in \mathcal{D}_{U_i}$, for any $i\in[[d]]$. Then, using Lemma~\ref{lemma:covering_cartesian}, the covering number of $\mathcal{D}_P$ with respect to the $\dist_{P}$ metric satisfies the bound
\begin{align}
\covering(\mathcal{D}_{P}, \dist_{P}, \epsilon) \leq \Big( \frac{3}{\epsilon} \Big)^{\prod \limits_{i=1}^{d} r_i} \Big( \frac{3 \bar{n}}{\epsilon} \Big)^{\sum \limits_{i=1}^{d} s_i r_i}, \epsilon \in (0,1).
\label{eq:coveringCartesian_2}
\end{align}

\noindent To finally derive a bound on the covering number of $\Class$, recall that we use the metric based on the Frobenius norm, denoted by $\dist_{\mathcal{G}}$, in order to cover the set $\Class$. Further, recall the mapping $\Phi$ defined as
\begin{align*}
\Phi(\Score, U_1, U_2,\ldots,U_d) = \Score \times_1 U_1 \times_2 U_2 \times_3 \cdots \times_d U_d \text{,}
\end{align*}
where $(\Score, U_1, U_2,\ldots,U_d) \in \mathcal{D}_{P}$. From this definition, it follows that $\Phi:\mathcal{D}_{P} \rightarrow \Class$. Then, given $P^{(1)}, P^{(2)}\in \mathcal{D}_{P}$, from application of Lemma~\ref{lemma:lipschitz_constant}, it follows that
\begin{align}
\dist_{\mathcal{G}} (  \Phi(P^{(1)}), \Phi(P^{(2)}) ) \leq \corebound (d+1) \dist_{P}(P^{(1)}, P^{(2)}),
\label{eq:Lipschitzmapping_2}
\end{align}
which implies the mapping $\Phi$ is Lipschitz with a Lipschitz constant of $\corebound \, (d+1)$. Using \eqref{eq:Lipschitzmapping_2} with \eqref{eq:coveringCartesian_2} and Lemma~\ref{lemma:lipscitz_map}, the statement of this lemma follows.
\qed

\section{Proof of Lemma~\ref{lemma:covering_sphere}}\label{app:CoveringNumber_HS}
Define $\mathcal{D}_{\Scorehat} := \{ \frac{\Score}{\corebound}:  \Score \in  \mathcal{D}_{\Score} \}$, so that $\| \Scorehat \|_1 \leq 1$ for all $\Scorehat \in \mathcal{D}_{\Scorehat}$. By application of Lemma~\ref{lemma:covering_sphere_main}, we have
\begin{align*}
\covering(\mathcal{D}_{\Scorehat}, \normed_1, \epsilon) \leq \Big( \frac{3}{\epsilon} \Big)^{\prod \limits_{i=1}^{d} r_i},
\end{align*}
for $\epsilon \in (0,1)$. The bound on $\covering(\mathcal{D}_{\Score}, \normed_1, \epsilon)$ follows from a volume comparison argument between $\mathcal{D}_{\Score}$ and $\mathcal{D}_{\Scorehat}$. \qed

\section{Proof of Lemma~\ref{lemma:covering_matrices}}\label{app:CoveringNumber_matrix}
The set $\mathcal{D}_{U}$ can be expressed as the Cartesian product of the sets $\mathcal{D}_{U}^{(j)} :=  \{ x \in \mathbb{R}^{n} : \|x\|_0 \leq s, \|x\|_2 \leq 1 \}$, $j \in [[r]]$. For any $j \in  [[r]]$, since there are $n \choose s$ ways to choose the support of an $s\mhyphen$sparse vector, we have
\begin{align}
\covering(\mathcal{D}_{U}^{(j)}, \normed_2, \epsilon)
\leq {n \choose s} \Big( \frac{3}{\epsilon} \Big)^{s} \text{,}
\label{eq:sparseVector}
\end{align}
with the application of Lemma~\ref{lemma:covering_sphere_main}. Then, the covering number of $\mathcal{D}_{U}$ with respect to the metric $\dist_{U}$, for any $\epsilon \in (0,1)$, satisfies the bound
\begin{align*}
\covering(\mathcal{D}_{U}, \dist_{U}, \epsilon)
\; &\leqa \;
\prod \limits_{j=1}^{r} \covering(\mathcal{D}_{U}^{(j)}, \normed_2, \epsilon)
\; \leqb \; \bigg[ {n \choose s} \Big( \frac{3}{\epsilon} \Big)^{s} \, \bigg]^{r}
\leq \frac{n^{sr}}{(s!)^{r}} \bigg( \frac{3}{\epsilon} \bigg)^{sr}
\nonumber\\
& = \; \bigg( \frac{3n}{(s!)^{\frac{1}{s}} \epsilon} \bigg)^{sr}
\leq \bigg( \frac{3 n}{\epsilon} \bigg)^{sr},
\end{align*}
\noindent where (a) and (b) follow from Lemma~\ref{lemma:covering_cartesian} and \eqref{eq:sparseVector}, respectively. \qed

\section{Proof of Lemma~\ref{lemma:lipschitz_constant}}\label{app:lipschitz_constant}
Let $\Ga, \Gb \in \Class$ such that
\begin{align*}
& \Ga = \Sa \times_1 \U_{a,1} \times_2 \U_{a,2} \times_3 \cdots \times_d \U_{a,d} \text{ , and} \\
& \Gb = \Sb \times_1 \U_{b,1} \times_2 \U_{b,2} \times_3 \cdots \times_d \U_{b,d} \text{ ,}
\end{align*}
where $\Sa, \Sb \in \mathcal{D}_{\Score}$, and $\U_{a,i}, \U_{b,i} \in \mathcal{D}_{U_i}$, $i\in[[d]]$. Then, we have
\begin{align}
\dist_{\mathcal{G}} ( \Ga, \Gb )
= & \| \Ga - \Gb \|_F
\nonumber\\
= & \| \Sa \times_1 \U_{a,1} \times_2 \U_{a,2} \times_3 \cdots \times_d \U_{a,d}
- \Sb \times_1 \U_{b,1} \times_2 \U_{b,2} \times_3 \cdots \times_d \U_{b,d} \|_F
\nonumber\\
= &  \| \Sa \times_1 \U_{a,1} \times_2 \U_{a,2} \times_3 \cdots \times_d \U_{a,d}
\nonumber\\
& \pm \Sa \times_1 \U_{a,1} \times_2 \U_{a,2} \times_3 \cdots \times_{d-1} \U_{a,d-1} \times_d \U_{b,d}
\nonumber\\
& \pm \Sa \times_1 \U_{a,1} \times_2 \U_{a,2} \times_3 \cdots \times_{d-2} \U_{a,d-2} \times_{d-1} \U_{b,d-1} \times_d \U_{b,d}
\nonumber\\
& \pm \cdots \pm \Sa \times_1 \U_{b,1} \times_2 \U_{b,2} \times_3 \cdots \times_{d-1} \U_{b,d-1} \times_d \U_{b,d}
\nonumber\\
&  - \Sb \times_1 \U_{b,1} \times_2 \U_{b,2} \times_3 \cdots \times_{d-1} \U_{b,d-1} \times_d \U_{b,d} \|_F
\nonumber\\
\leq \; &
\| \Sa \times_1 \U_{a,1} \times_2 \U_{a,2} \times_3 \cdots \times_{d-1} \U_{a,d-1} \times_d (\U_{a,d} - \U_{b,d}) \|_F
\nonumber\\
& + \| \Sa \times_1 \U_{a,1} \times_2 \U_{a,2} \times_3 \cdots \times_{d-2} \U_{a,d-2} \times_{d-1} (\U_{a,d-1} - \U_{b,d-1}) \times_d \U_{b,d} \|_F
\nonumber\\
& + \cdots +
\| \Sa \times_1 (\U_{a,1} - \U_{b,1}) \times_2 \U_{b,2} \times_3 \cdots \times_{d-1} \U_{b,d-1} \times_d \U_{b,d} \|_F
\nonumber\\
&  + \| (\Sa - \Sb) \times_1 \U_{b,1} \times_2 \U_{b,2} \times_3 \cdots \times_{d-1} \U_{b,d-1} \times_d \U_{b,d} \|_F \text{,}
\label{eq:covering1}
\end{align}

\noindent where $\pm \mathbf{V}$ denotes $+ \mathbf{V} - \mathbf{V}$ for any tensor $\mathbf{V} \in  \mathbb{R} ^ {n_1 \times n_2 \times \dots \times n_d }$. Define $\termj := \| \Sa \times_1 \U_{a,1} \times_2 \cdots \times_{j-1} \U_{a,j-1} \times_j (\U_{a,j} - \U_{b,j}) \times_{j+1} \U_{b,j+1} \times_{j+2} \cdots \times_{d} \U_{b,d}  \|_F$. With this definition, \eqref{eq:covering1} can be re-written as
\begin{align}
\dist_{\mathcal{G}} ( \Ga, \Gb )  \leq \sum \limits_{j=1}^{d} \termj + \| (\Sa - \Sb) \times_1 \U_{b,1} \times_2 \U_{b,2} \times_3 \cdots \times_{d-1} \U_{b,d-1} \times_d \U_{b,d} \|_F \text{.}
\label{eq:covering2}
\end{align}

\noindent We will bound the first $d$ terms and the last term in \eqref{eq:covering2} separately. Beginning with any term from among the first $d$ terms in \eqref{eq:covering2}, for any $j\in [[d]]$, we have

\begin{align}
\termj^2
&= \| \Sa \times_1 \U_{a,1} \times_2 \cdots \times_{j-1} \U_{a,j-1} \times_j (\U_{a,j} - \U_{b,j}) \times_{j+1} \U_{b,j+1} \times_{j+2} \cdots \times_{d} \U_{b,d}  \|_F^2
\nonumber\\
& = \sum \limits_{i_1, i_2, \ldots, i_d}
\big[ \big(\Sa \times_1 \U_{a,1} \times_2 \cdots \times_{j-1} \U_{a,j-1} \times_j (\U_{a,j} - \U_{b,j}) \times_{j+1} \U_{b,j+1} \times_{j+2} \cdots
\nonumber\\
& \qquad\qquad\qquad \cdots \times_{d} \U_{b,d} \big) (i_1, i_2, \ldots, i_d) \big]^2
\nonumber\\
& = \sum \limits_{i_1, i_2, \ldots, i_d}
\,  \Big( \sum \limits_{k_1, k_2, \ldots, k_d}  \Sa (k_1, k_2, \ldots, k_d) \; \U_{a,1} (i_1, k_1) \;
\cdots \; \big(\U_{a,j} - \U_{b,j} \big)(i_j, k_j) \; \cdots \; \U_{b,d} (i_d, k_d)  \Big)
\nonumber\\
& \qquad \qquad \quad \;
\Big(  \sum \limits_{l_1, l_2, \ldots, l_d}   \Sa (l_1, l_2, \ldots, l_d) \; \U_{a,1} (i_1, l_1) \; \cdots \; \big(\U_{a,j} - \U_{b,j} \big)(i_j, l_j) \; \cdots \; \U_{b,d} (i_d, l_d)   \Big)
\nonumber\\
& = \sum \limits_{i_1, i_2, \ldots, i_d}
\, \sum \limits_{k_1, k_2, \ldots, k_d}
\, \sum \limits_{l_1, l_2, \ldots, l_d}
\Sa (k_1, k_2, \ldots, k_d) \; \Sa (l_1, l_2, \ldots, l_d) \;
\U_{a,1} (i_1, k_1) \; \U_{a,1} (i_1, l_1) \;
\nonumber\\
& \qquad \qquad \quad \; \cdots \;
\big(\U_{a,j} - \U_{b,j} \big)(i_j, k_j) \; \big(\U_{a,j} - \U_{b,j} \big)(i_j, l_j)
\; \cdots \;
\U_{b,d} (i_d, k_d) \; \U_{b,d} (i_d, l_d)
\nonumber\\
& = \sum \limits_{k_1, k_2, \ldots, k_d} \, \sum \limits_{l_1, l_2, \ldots, l_d}
 \Sa (k_1, k_2, \ldots, k_d) \; \Sb (l_1, l_2, \ldots, l_d)
 \sum \limits_{i_1}  \U_{a,1} (i_1, k_1) \; \U_{a,1} (i_1, l_1)
 \nonumber\\
& \qquad \cdots \; \sum \limits_{i_j}  \big(\U_{a,j} - \U_{b,j} \big)(i_j, k_j) \; \big(\U_{a,j} - \U_{b,j} \big)(i_j, l_j) \; \cdots \;
   \sum \limits_{i_d}  \U_{b,d} (i_d, k_d) \; \U_{b,d} (i_d, l_d)
\nonumber\\
& \leqa \sum \limits_{k_1, k_2, \ldots, k_d} \, \sum \limits_{l_1, l_2, \ldots, l_d}
 \Sa (k_1, k_2, \ldots, k_d) \; \Sa (l_1, l_2, \ldots, l_d)
\nonumber\\
& \qquad\qquad\qquad \Big( \sum \limits_{i_j}  \big(\U_{a,j} - \U_{b,j} \big)(i_j, k_j) \big(\U_{a,j} - \U_{b,j} \big)(i_j, l_j) \Big)
 \nonumber\\
& \leq \;  \| \U_{a,j} - \U_{b,j} \|_{1,2}^2
\sum \limits_{k_1, k_2, \ldots, k_d} \, \sum \limits_{l_1, l_2, \ldots, l_d}
 \Sa (k_1, k_2, \ldots, k_d) \; \Sa (l_1, l_2, \ldots, l_d)
  \nonumber\\
& \leq  \; \| \U_{a,j} - \U_{b,j} \|_{1,2}^2
 \, \| \Sa \|_1 \, \| \Sa \|_1
 \leq \; \| \U_{a,j} - \U_{b,j} \|_{1,2}^2  \; \corebound^2 \, \text{,}
 \label{eq:covering3}
\end{align}
\noindent where (a) follows since $\mathbf{u}^{\top}\mathbf{v}\leq1$ for any column vectors $u$ and $v$ such that $\| u \|_2 \leq 1$ and $\| v \|_2 \leq 1$. Similarly, to bound the last term in \eqref{eq:covering2}, note that
\begin{align*}
& \| \big( \Sa - \Sb \big) \times_1 \U_{b,1} \times_2 \U_{b,2} \times_3 \cdots \times_{d-1} \U_{b,d-1} \times_d \U_{b,d} \|_F^2
\nonumber\\
& = \sum \limits_{i_1, i_2, \ldots, i_d}
\, \sum \limits_{k_1, k_2, \ldots, k_d}  \big(\Sa - \Sb \big) (k_1, k_2, \ldots, k_d) \; \U_{b,1} (i_1, k_1) \;
\U_{b,2} (i_2, k_2) \; \cdots \; \U_{b,d} (i_d, k_d)
\nonumber\\
& \qquad \qquad \quad \; \sum \limits_{l_1, l_2, \ldots, l_d}  \big( \Sa - \Sb \big) (l_1, l_2, \ldots, l_d) \; \U_{b,1} (i_1, l_1) \; \U_{b,1} (i_2, l_2) \; \cdots \; \U_{b,d} (i_d, l_d)
\nonumber\\
& = \sum \limits_{k_1, k_2, \ldots, k_d} \; \sum \limits_{l_1, l_2, \ldots, l_d} \big(\Sa - \Sb \big) (k_1, k_2, \ldots, k_d) \; \big(\Sa - \Sb \big) (l_1, l_2, \ldots, l_d)
\nonumber\\
& \qquad  \sum \limits_{i_1}\U_{b,1} (i_1, k_1) \; \U_{b,1} (i_1, l_1) \;
\sum \limits_{i_2} \U_{b,2} (i_2, k_2) \; \U_{b,2} (i_2, l_2) \;
\cdots \; \sum \limits_{i_d} \U_{b,d} (i_d, k_d) \; \U_{b,d} (i_d, l_d)
\end{align*}
\begin{align}
& \leq  \sum \limits_{k_1, k_2, \ldots, k_d} \sum \limits_{l_1, l_2, \ldots, l_d} \big(\Sa - \Sb \big) (k_1, k_2, \ldots, k_d) \big(\Sa - \Sb \big) (l_1, l_2, \ldots, l_d)
\leq  \| \Sa - \Sb \|_1^2  \text{.}
\label{eq:covering4}
\end{align}

Finally, using \eqref{eq:covering2} with \eqref{eq:covering3} and \eqref{eq:covering4}, we obtain
\begin{align}
\dist_{\mathcal{G}} ( \Ga, \Gb )
& \leq \sum \limits_{j=1}^{d} \, \corebound \, \| \U_{a,j} - \U_{b,j} \|_{1,2} + \| \Sa - \Sb \|_1
\nonumber\\
& \leq (d+1) \, \corebound \, \max \Big\{
\max \limits_{j\in[[d]]}  \{ \| \U_{a,j} - \U_{b,j} \|_{1,2} \}, \frac{1}{\corebound} \, \| \Sa - \Sb \|_1 \Big\}
\nonumber\\
&= (d+1) \, \corebound \, \dist_{P}(P^{(1)}, P^{(2)}) \text{,}
\end{align}
\noindent which completes the proof of this lemma.
\qed

\section{Proof of Theorem~\ref{th:subGaussian}}\label{app:proof_subGaussian}
We employ Theorem~\ref{th:deviationbound} with Lemma~\ref{lemma:low_rank_sparse_covering} to obtain a probabilistic bound on the restricted isometry property constant in Definition~\ref{def:RIP}. Before we can employ Theorem~\ref{th:deviationbound}, we need to evaluate bounds on the quantities $d_F( \Mset )$, $d_{2\rightarrow 2}( \Mset )$, $d_{4}( \Mset )$, and $\gamma_2(\Mset, \normed_2 )$, which we obtain as follows. We obtain a bound on $d_F( \Mset )$ as
\begin{align}\label{eq:bound1}
d_F( \Mset ) = \sup \limits_{\M \in \Mset} \|\M\|_F
\eqa \sup \limits_{\Z \in \Class} \|\Z\|_F
\leqb \corebound,
\end{align}
where (a) follows from the definition of $\Mset$ and (b) follows from the definition of $\Class$. Next, to obtain a bound on $d_{2\rightarrow 2}( \Mset )$ and $d_{4}( \Mset )$, note that for any $\Z \in \Class$ we have
$$\Vz \Vz^\top= \frac{1}{m} \;\; \mathbb{I}_{m} \otimes \Zvec^\top \Zvec=\frac{\|\Zvec\|_2^2}{m} \;\; \mathbb{I}_{m}$$
which leads to
\begin{align}\label{eq:bound2}
d_{2\rightarrow 2}( \Mset ) = \sup \limits_{\M \in \Mset} \|\M\|_{\normed_2}
= \sup \limits_{\Z \in \Class} \frac{\|\Z\|_F}{\sqrt{m}}
\leq \frac{\corebound}{\sqrt{m}} \text{, and}
\end{align}
\begin{align}
d_{4}^4( \Mset ) &= \sup \limits_{\M \in \Mset} \|\M\|_{S_4}^4
=  \sup \limits_{\M \in \Mset}  tr \Big[ (\M^\top \M)^2  \Big]
=  \sup \limits_{\M \in \Mset} tr \Big[ ( \M \M^\top )^2 \Big]
\nonumber\\
&= \sup \limits_{\Z \in \Class}  \; tr \Big[ \Big(  \frac{\|\Z\|_F^2}{m} \;\; \mathbb{I}_{m} \Big)^2 \Big]
= \sup \limits_{\Z \in \Class}  \; \frac{\|\Z\|_F^4}{m^2} \; tr \Big[ \, \mathbb{I}_{m} \, \Big]
\; \leq \;  \frac{\corebound^4}{m}\label{eq:bound3}.
\end{align}
Finally, to obtain a bound on the Talagrand's $\gamma_2$-functional, we employ \eqref{eq:TalagrandBoundMain} with Lemma~\ref{lemma:low_rank_sparse_covering} to obtain a bound on $\gamma_2(\Mset, \normed_2)$ as

\begin{align}
\gamma_2(\Mset, \normed_2)
& \leq C \int_{0}^{d_{2\rightarrow 2}(\Mset)}
\; \sqrt{\log \covering (\Mset, \normed_2, u)}  \; du
\leq C \mathlarger{\int_{0}^{\frac{\corebound}{\sqrt{m}}} }
\; \sqrt{\log \covering (\Mset, \normed_2, u)}  \; du
\nonumber\\
& = \frac{C}{\sqrt{m}} \mathlarger{\int_{0}^{\corebound}}
\; \sqrt{\log \covering (\Mset, \|.\|_{F}, \tilde{u})}  \; d\tilde{u}
\nonumber\\
& \leqc
C \; \sqrt{\frac{\prod \limits_{i=1}^{d} r_i + \sum \limits_{i=1}^{d} s_i r_i}{m}} \mathlarger{ \int_{0}^{\corebound}}
\; \sqrt{  \log \bigg(\frac{3 \bar{n} \corebound (d+1)}{\tilde{u}} \bigg)  }  \; d\tilde{u}
\nonumber\\
& \leqd C \; \sqrt{\frac{\prod \limits_{i=1}^{d} r_i + \sum \limits_{i=1}^{d} s_i r_i}{m}} \mathlarger{ \int_{0}^{\corebound}}
\; \log \bigg(\frac{3 \bar{n} \corebound (d+1)}{\tilde{u}} \bigg)  \; d\tilde{u}
\nonumber\\
& = C \; \sqrt{\frac{\prod \limits_{i=1}^{d} r_i + \sum \limits_{i=1}^{d} s_i r_i}{m}}
\; \Big[ \corebound \log \big( 3 \bar{n} (d+1) \big) + \corebound \Big]
\nonumber\\
& = \widetilde{C} \; \sqrt{\frac{ \corebound^2 \bigg( \prod \limits_{i=1}^{d} r_i + \sum \limits_{i=1}^{d} s_i r_i \bigg)}{m}}
\; \log \big( 3 \bar{n} d \big) \text{,}
\label{eq:GammaBound}
\end{align}
where $\widetilde{C} > 0$, (c) follows from Lemma~\ref{lemma:low_rank_sparse_covering}, and (d) follows since $\sqrt{\log_b(x/a)} \leq \log_b(x/a)$ for $x/a \in \mathbb{R}^{+}$, $b \in \mathbb{R}^{+}$, $x \geq a \, b$. Now that we have evaluated bounds on $d_F( \Mset )$, $d_{2\rightarrow 2}( \Mset )$, $d_{4}( \Mset )$, and $\gamma_2(\Mset, \normed_2)$, we can evaluate the quantities $E$, $U$, and $V$ in Theorem~\ref{th:deviationbound}. Evaluating a bound on $E$, we get
\begin{align}
E &=
\gamma_2(\Mset, \normed_2)^2 +
\gamma_2(\Mset, \normed_2) d_F(\Mset)  + d_F(\Mset) d_{2 \rightarrow 2}(\Mset)
\nonumber\\
& \leqe
\widetilde{C}^2 \; \frac{ \corebound^2 \bigg( \prod \limits_{i=1}^{d} r_i + \sum \limits_{i=1}^{d} s_i r_i \bigg)}{m} \; \Big( \log \big( 3 \bar{n} d \big) \Big)^2
+ \widetilde{C} \, \corebound \; \sqrt{\frac{ \corebound^2 \bigg( \prod \limits_{i=1}^{d} r_i + \sum \limits_{i=1}^{d} s_i r_i \bigg) \; \log \big( 3 \bar{n} d \big) }{m}}
+ \frac{\corebound^2 }{\sqrt{m}}
\nonumber\\
& \leqf \;
\frac{\delta^2 \, \widetilde{C}^2}{K_1} + \frac{\corebound \, \delta \, \widetilde{C}}{\sqrt{K_1}}
+ \frac{ \corebound \, \delta}{\sqrt{K_1}}
\; \; \leqg \; \;
\frac{\delta \, \widetilde{C}^2}{K_1} + \frac{\corebound \, \delta \, \widetilde{C}}{\sqrt{K_1}}
+ \frac{ \corebound \, \delta}{\sqrt{K_1}}
\;\; \leq \;\;
\frac{\delta \, \big( \widetilde{C}^2 + \widetilde{C} \, \corebound + \corebound \big)}{ \min \{K_1, \sqrt{K_1} \} } \text{,}
\label{eq:Ebound1}
\end{align}
where (e) follows from application of \eqref{eq:bound1} and \eqref{eq:bound2} with \eqref{eq:GammaBound}, (f) follows from the bound on $m$, and (g) follows since $\delta \in (0,1)$. Setting $K_1 \geq \max \bigg\{ \Big( 2 c_1 (\widetilde{C}^2 + \widetilde{C} \, \corebound + \corebound)  \Big)^2, 2 c_1 (\widetilde{C}^2 + \widetilde{C} \, \corebound + \corebound) \bigg\}$ in \eqref{eq:Ebound1} for some $c_1 > 0$, we obtain
\begin{align}
c_1 E
\leq
\frac{\delta \, c_1 \,  \big( \widetilde{C}^2 + \widetilde{C} \, \corebound + \corebound \big)}{ \min \{K_1, \sqrt{K_1} \} }
\leq \frac{\delta}{2}.
\label{eq:Ebound2}
\end{align}
Next, we can evaluate bounds on $U$ and $V$ as
\begin{align}
U &= d_{2 \rightarrow 2}^2(\Mset) \leqh \frac{\corebound^2}{m} \text{, and  }
V = d_{4}^2(\Mset) \leqi \frac{\corebound^2}{\sqrt{m}} \text{,}
\end{align}
where (h) follows from \eqref{eq:bound2} and (i) follows from \eqref{eq:bound3}. Finally, we use these bounds on $U$ and $V$ to bound the quantity $2 \exp \big( - c_2 \min \{ \frac{t^2}{V^2}, \frac{t}{U} \}  \big)$as
\begin{align}
2 \exp \Big( - c_2 \min \Big\{ \frac{t^2}{V^2}, \frac{t}{U} \Big\}  \Big)
&\leq 2 \exp \Big( - c_2 \min \Big\{ m \, \Big( \frac{t}{\corebound^2} \Big)^2, \frac{tm}{\corebound^2} \Big\}  \Big)
\nonumber\\
&\leqj 2 \exp \Big( - c_2 \min \Big\{ \Big( \frac{\delta}{2 \corebound^2} \Big)^2 \frac{K_2 \, \log (\varepsilon^{-1})}{\delta^2}, \frac{K_2 \, \log (\varepsilon^{-1})}{2 \corebound^2 \, \delta} \Big\}  \Big)
\nonumber\\
& = 2 \exp \Big( - \frac{c_2 \, K_2 \, \log (\varepsilon^{-1})}{2 \delta}
\min \Big\{ \frac{\delta}{2 \corebound^4} , \frac{1}{\corebound^2} \Big\}  \Big)
\leqk \varepsilon \text{,}
\label{eq:prob_bound}
\end{align}
where (j) follows from setting $t = \frac{\delta}{2}$ and using the bound on $m$, while (k) holds true for
$$K_2 \geq \max \Big\{ (2 \corebound^2)^2  , 2 \delta \corebound^2 \Big\} \Big( \frac{\log (1/2)}{c_2 \log(\varepsilon)} + \frac{1}{c_2} \Big) \text{.}$$
Using \eqref{eq:Ebound2}, \eqref{eq:prob_bound}, and $t = \frac{\delta}{2}$ with Theorem~\ref{th:deviationbound}, the proof of this theorem follows.
\qed

\section{Auxiliary Lemmas}\label{app:auxiliary}
\begin{lemma}\label{lemma:covering_sphere_main} [\cite{candes2011tight}]
For any fixed notion of norm $\normed$, define a unit-norm ball $\mathcal{B}_{1} := \{ x \in \mathbb{R}^{n} : \|x\| \leq 1  \}$ with distance measure $\normed$. Then the covering number of $\mathcal{B}_{1}$ (with respect to the norm $\normed$) satisfies the bound
\begin{align*}
\covering (\mathcal{B}_{1}, \normed, \epsilon) \leq \Big( \frac{3}{\epsilon} \Big)^{n}, \epsilon \in (0,1).
\end{align*}
\end{lemma}
\begin{lemma} [\cite{gribonval2015sample}]
Define metric spaces $(\mathcal{D}_{1}, h_1$, $(\mathcal{D}_{2}, h_2)$, $\ldots$, $(\mathcal{D}_p, h_p)$. Further, define the Cartesian product $\mathcal{D}_0 := \mathcal{D}_1 \times_1 \mathcal{D}_2 \times_2 \cdots \times_p \mathcal{D}_p$ with respect to the norm $h_0( D_0^{1}, D_0^{2} ) =  \max_{j\in [[p]]} \big\{ h_j( D_j^{1}, D_j^{2} ) \big\}$, where $D_0^{1}, D_0^{2} \in \mathcal{D}_0$ such that $D_0^{1} = D_1^{1} \times_1 D_2^{1} \times_2 \cdots \times_p D_p^{1}\,$, $D_0^{2} = D_1^{2} \times_1 D_2^{2} \times_2 \cdots \times_p D_p^{2}\,$, and $D_j^{1}, D_j^{2} \in \mathcal{D}_j$ for any $j\in [[p]]$. Then,  $\covering (\mathcal{D}_{0}, h_0, \epsilon)$ satisfies the bound $\covering (\mathcal{D}_{0}, h_0, \epsilon) \leq \prod \limits_{j=1}^{d} \covering (\mathcal{D}_{j}, h_j, \epsilon). $
\label{lemma:covering_cartesian}
\end{lemma}

\begin{lemma} [\cite{szarek1997metric}]
Define sets $\mathcal{D}_{1}$ and $\mathcal{D}_{2}$ with distance measures $h_1$ and $h_2$, respectively. Further, define map $\Phi : \mathcal{K} \rightarrow \mathcal{D}_2$ such that $\mathcal{K} \subset \mathcal{D}_1$. Then, for some $L>0$, if $\Phi$ satisfies
\begin{align*}
h_2 (\Phi(K_1), \Phi(K_2) ) \leq \, L \, h_1(K_1, K_2) \;\; \text{for } K_1, K_2 \in \mathcal{K} \text{,}
\end{align*}
i.e. $\Phi$ is a Lipschitz map with constant $L$, then, for any $\epsilon>0$, we have
\begin{align*}
\covering (\Phi( \mathcal{K} ), h_2, L \, \epsilon) \leq \covering (\mathcal{K}, h_1, \epsilon) \text{.}
\end{align*}
\label{lemma:lipscitz_map}
\end{lemma}

\end{document}